\documentclass{article}

\usepackage{microtype}
\usepackage{graphicx}
\usepackage{subcaption}
\usepackage{booktabs} 

\usepackage{hyperref}

\usepackage[accepted]{icml2026}

\usepackage{amsmath}
\usepackage{amssymb}
\usepackage{mathtools}
\usepackage{amsthm}

\usepackage[capitalize,noabbrev]{cleveref}

\theoremstyle{plain}

\theoremstyle{definition}

\theoremstyle{remark}

\usepackage[textsize=tiny]{todonotes}

\usepackage{color}
\usepackage[dvipsnames]{xcolor}
\usepackage{colortbl}

\definecolor{Gray}{gray}{0.9}
\definecolor{LightBlue}{RGB}{236,244,249}
\definecolor{darkorange}{rgb}{1.0, 0.55, 0.0}

\definecolor{color1}{HTML}{ECF4F9}
\definecolor{color2}{HTML}{FFF1E0}
\definecolor{color3}{HTML}{ECF4E9}
\newcommand{\CC}[1]{\cellcolor{color1}}
\newcommand{\RC}[1]{\rowcolor{color1}}
\newcommand{\RD}[1]{\rowcolor{color2}}
\newcommand{\RE}[1]{\rowcolor{color3}}
\newcommand{\GR}[1]{\rowcolor{gray!#1}}

\usepackage{multirow}
\usepackage[normalem]{ulem}
\useunder{\uline}{\ul}{}
\usepackage{xspace}
\newcommand{\Method}{\texttt{VideoTemp-o3}\xspace}

\makeatletter
\DeclareRobustCommand\onedot{\futurelet\@let@token\@onedot}
\def\@onedot{\ifx\@let@token.\else.\null\fi\xspace}

\def\eg{\emph{e.g}\onedot} 
\def\ie{\emph{i.e}\onedot}

\makeatother

\icmltitlerunning{VideoTemp-o3: Harmonizing Temporal Grounding and Video Understanding in Agentic Thinking-with-Videos}

\begin{document}

\twocolumn[
  \icmltitle{VideoTemp-o3: Harmonizing Temporal Grounding and Video Understanding \\ in Agentic Thinking-with-Videos}



  \icmlsetsymbol{equal}{*}
  \icmlsetsymbol{corresponding}{$\dagger$}
  \icmlsetsymbol{projectleader}{$\ddagger$}

  \begin{icmlauthorlist}
    \icmlauthor{Wenqi Liu}{sdu,equal}
    \icmlauthor{Yunxiao Wang}{sdu,equal}
    \icmlauthor{Shijie Ma}{casia,equal}
    \icmlauthor{Meng Liu}{sdu,corresponding}
    \icmlauthor{Qile Su}{beihang} \\
    \icmlauthor{Tianke Zhang}{kuaishou}
    \icmlauthor{Haonan Fan}{kuaishou}
    \icmlauthor{Changyi Liu}{kuaishou}
    \icmlauthor{Kaiyu Jiang}{kuaishou}
    \icmlauthor{Jiankang Chen}{kuaishou}
    \icmlauthor{Kaiyu Tang}{kuaishou} \\
    \icmlauthor{Bin Wen}{kuaishou,projectleader}
    \icmlauthor{Fan Yang}{kuaishou}
    \icmlauthor{Tingting Gao}{kuaishou}
    \icmlauthor{Han Li}{kuaishou}
    \icmlauthor{Yinwei Wei}{sdu}
    \icmlauthor{Xuemeng Song}{sustech,corresponding}
    \vskip 0.1in
    \texttt{\textcolor{blue}{\href{https://liuwq-bit.github.io/VideoTemp-o3}{https://liuwq-bit.github.io/VideoTemp-o3}}}
  \end{icmlauthorlist}

  \icmlaffiliation{sdu}{Shandong University}
  \icmlaffiliation{casia}{Institute of Automation, Chinese Academy of Sciences}
  \icmlaffiliation{beihang}{Beihang University}
  \icmlaffiliation{kuaishou}{Kuaishou Technology}
  \icmlaffiliation{sustech}{Southern University of Science and Technology}

  \icmlcorrespondingauthor{Meng Liu}{mengliu.sdu@gmail.com}
  \icmlcorrespondingauthor{Xuemeng Song}{sxmustc@gmail.com}



  \vskip 0.2in
]



\printAffiliationsAndNotice{\icmlEqualContribution\icmlProjectLeader}

\begin{abstract}
In long-video understanding, conventional uniform frame sampling often fails to capture key visual evidence, leading to degraded performance and increased hallucinations. To address this, recent agentic \emph{thinking-with-videos} paradigms have emerged, adopting a \emph{localize–clip–answer} pipeline in which the model actively identifies relevant video segments, performs dense sampling within those clips, and then produces answers. However, existing methods remain inefficient, suffer from weak localization, and adhere to rigid workflows. To solve these issues, we propose \Method, a unified agentic \emph{thinking-with-videos} framework that jointly models video grounding and question answering. \Method exhibits strong localization capability, supports on-demand clipping, and can refine inaccurate localizations.  Specifically, in the supervised fine-tuning stage, we design a unified masking mechanism that encourages exploration while preventing noise. For reinforcement learning, we introduce dedicated rewards to mitigate reward hacking. Besides, from the data perspective, we develop an effective pipeline to construct high-quality long video grounded QA data, along with a corresponding benchmark for systematic evaluation across various video durations. Experimental results demonstrate that our method achieves remarkable performance on both long video understanding and grounding. 
\end{abstract}
\section{Introduction}
\label{sec:intro}

Although Multimodal Large Language Models (MLLMs)~\cite{bai2025qwen2,team2025kwai,team2025kimi} have made remarkable advances, their internal knowledge remains static once trained, limiting their capacity to tackle complex tasks in dynamic environments. To address this issue, recent works introduce agentic MLLMs capable of invoking external tools. For example, when encountering up-to-date and knowledge-intensive questions, agentic models can proactively invoke search tools~\cite{wu2025mmsearch,narayan2025deepmmsearch}. For fine-grained perception~\cite{tong2024eyes,ma2025genhancer} of high-resolution inputs, OpenAI-o3 has pioneered the \emph{thinking-with-images} paradigm~\cite{zheng2025deepeyes,zhang2025thyme,hong2025deepeyesv2}, enabling image transformations that amplify salient details while suppressing irrelevant content.

In long-video understanding~\cite{feng2025video,li2025videochat}, conventional approaches typically rely on uniform frame sampling under a fixed frame budget to control computational cost. Such a rigid pipeline yields sparse visual evidence and often overlooks query-relevant content, leading to substantial performance degradation. Therefore, taking the spirit of \emph{thinking-with-images}, \emph{thinking-with-videos} has emerged to enable query-aware fine-grained temporal grounding~\cite{liu2018attentive, mun2020local, wang2022siamese, lin2023univtg, liu2023survey, wang2026time}. In this paradigm, MLLMs first proactively identify the video segments most pertinent to the question. They then perform dense sampling within these intervals to produce reliable, clip-grounded responses.
In the literature, VideoExplorer~\cite{yuan2025videoexplorer} proposes a multi-agent framework composed of a planner, a grounder, and an understanding model, which collaboratively achieve grounded video understanding. Later works like VITAL~\cite{zhang2025thinking}, REVISOR~\cite{li2025revisor}, and LongVT~\cite{yang2025longvt} adopt hybrid training that combines Supervised Fine-Tuning (SFT) and Reinforcement Learning (RL) on curated datasets, enabling models to effectively identify key segments and produce accurate responses.

\begin{figure*}[!t]
  \centering
  \includegraphics[width=.93\textwidth]{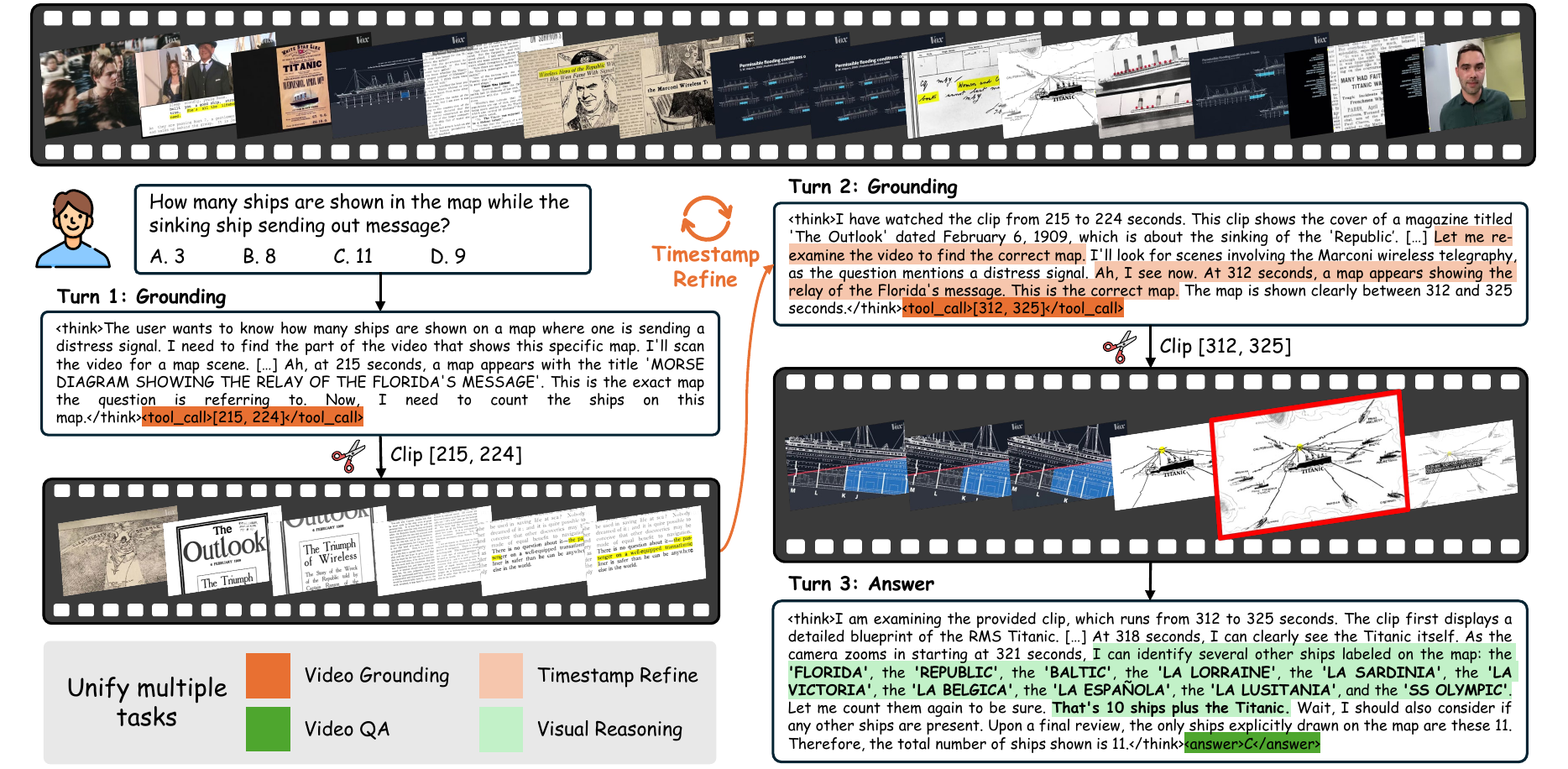}
  \caption{Illustration of the agentic pipeline in \Method. Given the video QA pair, it performs on-demand grounding and refines the initial rough segment. Finally, it produces a reliable answer grounded in the pertinent visual evidence.}
  \vspace{-5pt}
  \label{fig:main}
\end{figure*}

Despite these advancements, several critical limitations persist: (1) \emph{Workflow complexity.} Many existing methods rely on multiple specialized models to separately perform temporal grounding and video question answering, incurring substantial inference overhead. (2) \emph{Imprecise grounding.} Many approaches struggle with precise grounding and offer limited mechanisms to evaluate or refine grounding results. (3) \emph{Rigid pipelines.} Most methods follow a rigid procedure that the model blindly crops the video \emph{once} and then immediately produces an answer, leading to unnecessary computation for short videos that could be answered directly, while lacking support for iterative grounding refinement in challenging long-video scenarios.
We attribute these limitations to three primary factors: (1) \emph{Suboptimal training strategies.} Existing SFT supervision and RL reward designs are insufficient to promote precise grounding or support iterative refinement of inaccurate intervals. (2) \emph{Low-quality data.} Current grounding annotations are generally shifted and are dominated by short videos. (3) \emph{Lack of instructional trajectories to incentivize desirable patterns.} There is a lack of high-quality, multi-turn trajectories that exemplify \emph{localize-clip-answer} behavior, hindering MLLMs from effectively internalizing the \emph{thinking-with-videos} patterns.

\begin{figure*}[!t]
  \centering
  \includegraphics[width=.8\textwidth]{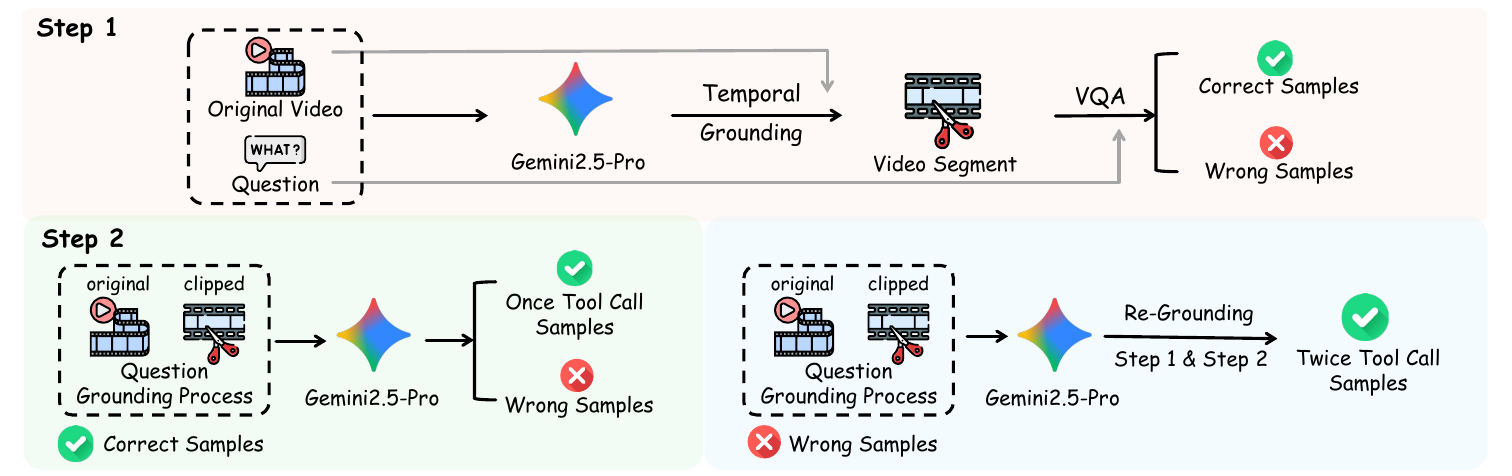}
  \caption{Multi-turn, multi-tool call data curation pipeline.}
  \vspace{-10pt}
  \label{fig:data_pipeline}
\end{figure*}

To address these challenges, we propose \Method, a unified \emph{thinking-with-videos} framework that jointly integrates video question answering (VideoQA) and temporal grounding within a single model. \Method supports on-demand grounding and cropping, enabling iterative refinement until sufficient evidence is gathered to produce a reliable answer, as in Fig.~\ref{fig:main}. Specifically, we begin by establishing a foundational \emph{localize-clip-answer} capability through cold-start SFT. To facilitate learning across both VideoQA and grounding, we introduce a unified masking strategy for multi-turn supervision that encourages exploration while preserving reliable learning signals. In the subsequent RL stage, we design tailored rewards to significantly improve grounding precision and clip-grounded answer accuracy, while effectively mitigating reward hacking. In this way, \Method strengthens its internal grounding ability, which is the prerequisite for \emph{thinking-with-videos} paradigm. From the data perspective, we design a dedicated pipeline to construct large-scale long-video grounded QA (GQA) datasets with accurate temporal segments and corresponding answers, ensuring strong alignment between grounded evidence and responses. Finally, we introduce VideoTemp-Bench, a benchmark for evaluating GQA across videos of diverse durations, as well as in-depth analyses.

In summary, our main contributions are as follows:
\begin{itemize}
    \vspace{-5pt}
    \item We present \Method, an agentic \emph{thinking-with-videos} model that harmonizes temporal grounding and VideoQA within a single architecture. It supports on-demand video cropping and enables multi-turn grounding refinement.
    \vspace{-5pt}
    \item We develop a cold-start SFT strategy and tailor-made RL rewards to enhance the model's internal grounding performance. The training paradigm effectively instills \emph{thinking-with-videos} behavior and yields stronger video understanding.
    \vspace{-5pt}
    \item We introduce a high-quality pipeline to curate a large-scale long-video GQA dataset. The resulting data tightly aligns cropped segments with answers.
    \vspace{-5pt}
    \item Extensive experiments demonstrate that \Method achieves state-of-the-art performance across several video understanding benchmarks. We also present VideoTemp-Bench that highlights the limitations of current models and provides in-depth analyses.
\end{itemize}


\section{Related Work}
\label{sec:related_work}

\paragraph{Agentic Multimodal Large Language Models.}
Agentic MLLMs substantially improve real-world problem-solving by leveraging external tools. In particular, for up-to-date or information-seeking questions beyond the model's internal knowledge, agentic search models~\cite{wu2025mmsearch,narayan2025deepmmsearch} can effectively identify the boundaries of their capabilities. When needed, they autonomously invoke search engines to retrieve external information, enabling more accurate responses and reducing hallucinations.
In scenarios requiring fine-grained perception~\cite{tong2024eyes, ma2024active, ma2025protogcd}, such as those involving small foreground objects, OpenAI-o3 introduced the \emph{thinking-with-images} paradigm~\cite{zheng2025deepeyes}, which enables models to invoke external tools for operations like cropping and zooming to better extract critical visual cues, and significantly enhances fine-grained understanding. Inspired by this, Thyme~\cite{zhang2025thyme} and DeepEyesV2~\cite{hong2025deepeyesv2} leverage code generation as a universal interface for flexible image manipulation.
Recently, the \emph{thinking-with-videos} paradigm has been proposed, extending fine-grained perception of \emph{thinking-with-images} from the spatial dimension to the temporal dimension.

\vspace{-12pt}
\paragraph{Long Video Understanding.}
Recent works like VideoChat-R1~\cite{li2025videochat} and Video-R1~\cite{feng2025video} have adopted RL-based strategies to improve general video understanding and reasoning. However, long video understanding remains challenging. A key bottleneck is the prevalent use of uniform sampling under a fixed frame budget, introduced to control computational cost, which often misses critical visual evidence.
To alleviate this, LongVA~\cite{zhang2024long} and LongVILA~\cite{chen2024longvila} employ long-context fine-tuning to extend context length for long videos. Others~\cite{li2024llama,hyun2025multi,wang2025adaretake} resort to visual token compression.
More recently, the agentic \emph{thinking-with-videos} paradigm has emerged as a complementary approach. Some works, including VideoChat-R1.5~\cite{yan2026videochat}, VideoThinker~\cite{wang2025video}, Open-o3 Video~\cite{meng2025open}, and Rewatch-R1~\cite{zhang2026rewatchr}, can output timestamps or bounding boxes during reasoning, but they mainly follow a text-only CoT paradigm without explicit tool calls, and therefore provide limited support for multi-round localization and answering. Another line of work first localizes or selects relevant visual content and then answers based on the selected evidence. For example, VideoExplorer~\cite{yuan2025videoexplorer} leverages multi-agent collaboration, with specialized modules handling planning, grounding, and understanding. VITAL~\cite{zhang2025thinking} adopts a two-stage training process involving SFT-RL to elicit the model's segment cropping. REVISOR~\cite{li2025revisor} introduces a two-stage inference paradigm that first grounds the relevant intervals and then generates responses based on the selected content. Similarly, LongVT~\cite{yang2025longvt} proposes a three-stage training SFT-RL-RFT strategy to progressively enhance \emph{thinking-with-videos} behavior. FrameThinker~\cite{he2026framethinker} also performs multi-turn visual selection before answering by spotlighting informative frames. These methods demonstrate the effectiveness of focusing on key visual evidence, while our work further unifies temporal grounding and VideoQA in a single model with on-demand cropping, iterative refinement, and dedicated training designs for noisy multi-turn trajectories.

\section{Dataset and Benchmark}
\label{sec:db}

\subsection{Data Construction Pipeline}
Existing datasets exhibit several limitations that hinder the induction of \emph{thinking-with-videos} behavior. (1) Grounding annotations are often low-quality, with coarse and temporally misaligned intervals. (2) Long video QA with explicit temporal grounding annotations is scarce. (3) Since accurate grounding in a single pass is sometimes unrealistic in long videos, MLLMs must learn reflection, \ie, refining grounding results through multi-turn cropping. To address these limitations, we propose a dedicated pipeline for constructing both single-turn and multi-turn data.

\vspace{-12pt}
\paragraph{Single-turn Data without Tool Call.}
To establish fundamental reasoning ability, we construct high-quality Chain-of-Thought (CoT) training samples for both QA and grounding.
We employ a rejection sampling strategy to ensure reliable reasoning chains. Specifically, for the VideoQA data, we utilize Qwen3-VL-235B-A22B-Thinking~\cite{bai2025qwen2, qwen3technicalreport} to generate both the reasoning trajectories and final answers. Only samples whose predicted answers match the ground truth are retained.
For grounding, we re-annotate and filter data to improve temporal precision. Concretely, we leverage Gemini-2.5-Pro~\cite{comanici2025gemini} to re-localize key segments and keep samples whose predicted intervals achieve IoU$>$0.5 with the original annotations, ensuring precise and reliable grounding labels for training.

\vspace{-12pt}
\paragraph{Multi-turn Data with Tool Call.}
The core objective of multi-turn data is to simulate realistic, tool-assisted grounding behaviors. By iteratively identifying, cropping, and validating relevant video segments for a given QA task, the framework enables the construction of high-quality multi-turn training samples that explicitly involve reflection and verification of previously localized content. A key component of our design is a dedicated verification stage, which critically evaluates whether the selected video segments provide sufficient evidence to support a coherent \emph{thinking-with-videos} reasoning chain.

As shown in Fig.~\ref{fig:data_pipeline}, in Step~1, given an original video and its question, we first prompt Gemini-2.5-Pro~\cite{comanici2025gemini} to predict a candidate temporal segment. We crop this segment and feed it into the model along with the original question, while explicitly constraining the model to answer the question solely based on the segment. Only samples whose predicted answers match the ground truth are retained. This filtering step ensures that the localized video segment contains sufficient information to correctly answer the question.
In Step~2, we perform a closed-loop consistency verification. Specifically, we provide the model with the full context, including original video, question, grounding process and cropped segment. We then instruct the model to reason based on the localized segment and prior context. Samples are retained only if the final answer remains consistent with the ground-truth, forming multi-turn, single-tool-call training data.
For samples that fail either verification stage, we prompt Gemini-2.5-Pro to re-ground the video by leveraging the accumulated context from the previous failure attempt, and then repeat both step 1 and step 2. Samples that pass after one additional refinement round are retained as multi-turn, multi-tool-call training data.
To ensure that iterative grounding is truly necessary, we apply this procedure only to long videos ($>$3min).
For questions with multiple possible evidence segments, this verification process ensures that the selected segment is sufficient to answer the question, even if other valid evidence also exists elsewhere in the video.

\begin{figure}[!t]
  \centering
  \begin{subfigure}{\columnwidth}
    \centering
    \includegraphics[width=.95\linewidth]{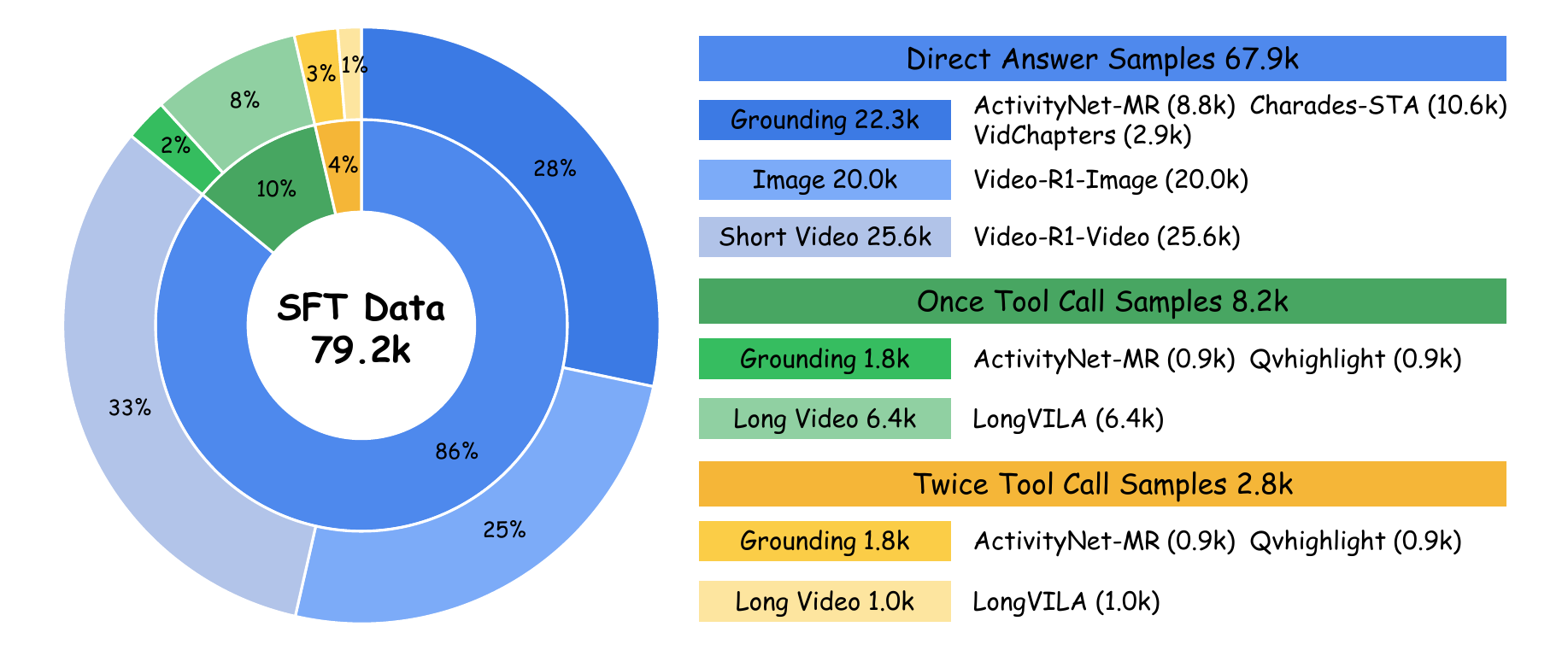}
    \vspace{-5pt}
    \caption{Distribution of SFT Data.}
    \label{fig:sft_data_dist}
  \end{subfigure}
  \begin{subfigure}{\columnwidth}
    \centering
    \includegraphics[width=.95\linewidth]{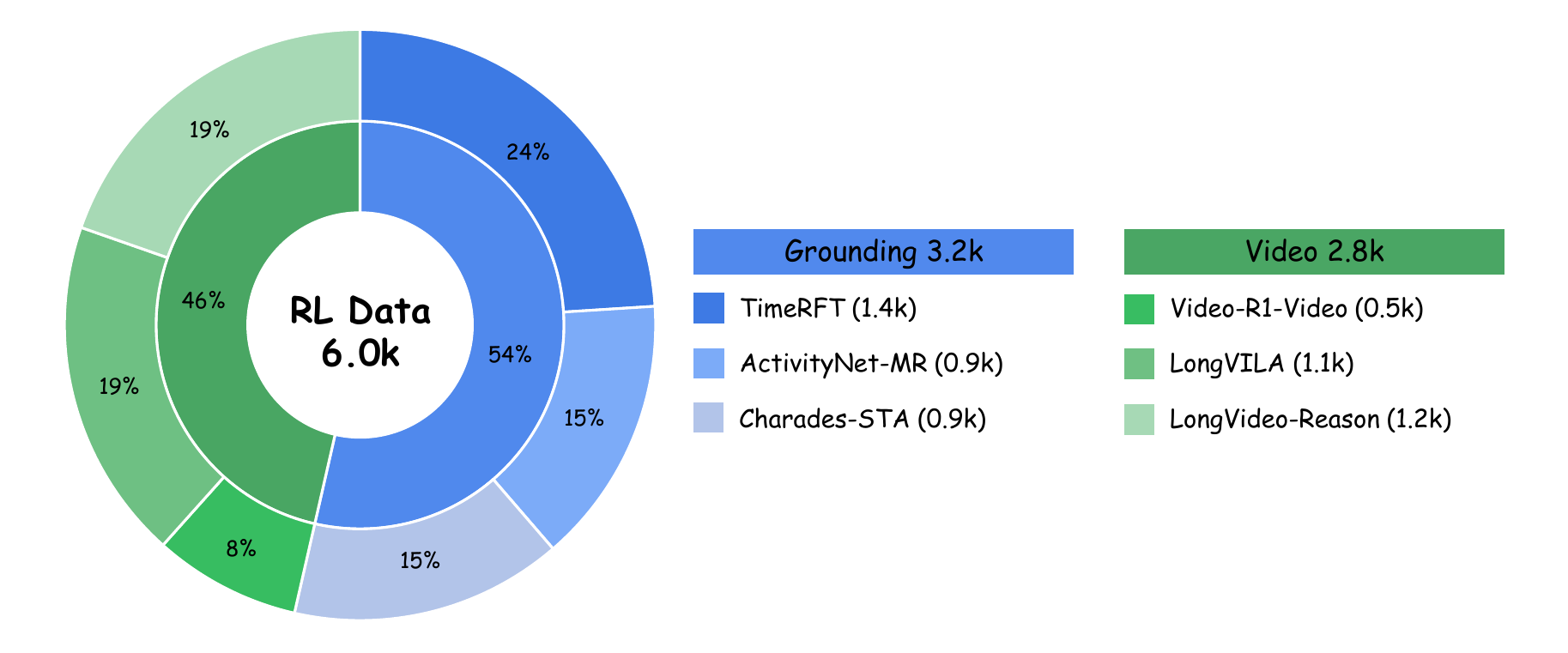}
    \vspace{-5pt}
    \caption{Distribution of RL Data.}
    \label{fig:rl_data_dist}
  \end{subfigure}
  \caption{Training Data Distribution.}
  \label{fig:training_data_dist}
  \vspace{-10pt}
\end{figure}

\subsection{Training Data}
\label{subsec:train-data}

\paragraph{SFT Data.}
To incentivize the \emph{thinking-with-videos} behavior, we carefully curated the task distribution and data sources for the SFT dataset. It encompasses both single- and multi-turn trajectories, spans grounding and QA tasks, with diverse video durations. The detailed composition is shown in Fig.~\ref{fig:sft_data_dist}.
For data formatting, the reasoning path is enclosed in $\texttt{<think>...</think>}$ followed by tool call parameters in $\texttt{<tool\_call>...</tool\_call>}$, while final answer outputs adopt the format $\texttt{<answer>...</answer>}$.

\vspace{-10pt}
\paragraph{RL Data.}
To further enhance the model’s capabilities in video grounding and videoQA, we employ a mixture of two data types during the RL stage. These data sources are designed to comprehensively cover the key dimensions of grounding and video reasoning. The detailed composition is shown in Fig.~\ref{fig:rl_data_dist}.
To ensure reliable reward signals for RL, all data annotations undergo rigorous human verification.
For temporal grounding, low-quality samples are initially filtered, and the remainder undergoes careful human verification and correction. For long-video grounded QA data with both interval and answer annotations, annotators carefully inspect both, discarding ambiguous cases and correcting annotation errors to maintain high data fidelity. When a question can be answered from multiple temporal segments, annotators mark all valid evidence intervals whenever possible, so that alternative correct localizations are not unfairly penalized during RL.

\vspace{-5pt}
\paragraph{Data Balance.}
To maintain generalization across diverse scenarios, we employ careful data balancing during training from three dimensions: (1) \emph{Modality Balance}. We carefully regulate the proportion of video data while simultaneously maintaining some image data, which preserves general multimodal reasoning.
(2) \emph{Task Balance}. A diverse and balanced set of question types is included, \eg, temporal grounding, temporal understanding, entity/action recognition, and relational reasoning.
(3) \emph{Video-length Balance}. We perform stratified sampling across a range of video durations. More details are presented in the Appendix~\ref{appendix:dataset_and_benchmark}.

\subsection{VideoTemp-Bench}
\label{subsec:videotemp-bench}
To thoroughly examine the interaction between video grounding and understanding abilities, we conduct evaluations on video GQA datasets. However, existing video GQA benchmarks predominantly target short videos ($<$3min).
To enable systematic evaluation of the model’s GQA performance across varying temporal scales, we introduce VideoTemp-Bench. This benchmark segments videos into four duration-based categories: 0$\sim$3 minutes, 3$\sim$10 minutes, 10$\sim$20 minutes, and $>$20 minutes, with 300 samples in each category and 1,200 samples in total. It is designed to reveal how video understanding performance, \ie, both temporal localization and final answering, varies as video duration increases. Please refer to the Appendix~\ref{appendix:dataset_and_benchmark}.

\section{Method}
\label{sec:method}

\paragraph{Overview.}
We introduce our agentic \emph{thinking-with-videos} model, namely \Method. We first present task formulation. Then we introduce training strategies including SFT in Sec.~\ref{subsec:method-sft} and RL in Sec.~\ref{subsec:method-rl}.


\vspace{-10pt}
\paragraph{Task Formulation.}
We treat \emph{thinking-with-videos} as a \emph{localize-crop-answer} process. For long videos where uniform sampling often fails to capture key evidence, agentic MLLMs can iteratively invoke tools, \ie, video cropping and dense sampling, to retrieve relevant frames and then answer based on them.
In this way, we can bypass the sparse evidence challenge in uniform sampling.
Formally, given a video-question pair $(V,Q)$, the model first skims the video at a low sampling rate $s_0$. Then it engages in iterative interaction turns. At each turn, it generates a textual reasoning $T$, along with either a temporal interval $P$ or a final answer $A$. If $P$ is predicted, an external cropping module extracts the corresponding clip from $V$ at a higher rate $s_d>s_0$, \ie, $C=\text{Crop}(V,P,s_d)$. The cropped clip $C$ is appended to the context for the next turn. The interaction terminates either when the model outputs the answer $A$ or reaches the maximum number of turns $T_{max}$.
As a result, each training sample $i$ can be represented as a multi-turn trajectory:
\begin{equation}
    \tau_i =\{(V,Q);([T_{i,1},P_{i,1}, C_{i,1}], \cdots, [T_{i,t}, A_i])\},
    \label{eq:traj}
\end{equation}
where $t$ denotes the index of the final turn of this trajectory.

\subsection{Cold Start Supervised Fine-Tuning}
\label{subsec:method-sft}

We begin by cold-start SFT to incentivize \emph{thinking-with-videos} behaviors, where the model learns to first perform on-demand crop, then produce the final answer grounded in key visual evidence in the segments.

\vspace{-10pt}
\paragraph{Three Desirable Features.}
To enable the model to solve complex problems flexibly and effectively, we summarize the following properties:
(1) \emph{On-demand cropping.} For short videos, the model can efficiently answer directly without cropping, \ie, $t=0$ in Eq.~\eqref{eq:traj}. As a result, we include short VideoQA data without any cropping.
(2) \emph{Reflection mechanism.} For challenging long videos, the initial grounding $P_1$ may be inaccurate, the model could perform multiple refinement rounds, \ie, $t \ge 3$. In this regard, we curate multi-tool-call training samples with grounding reflection trajectories.
(3) \emph{Unifying temporal grounding and VideoQA.} Our model supports both VideoQA and grounding, so the answer $A_i$ in Eq.~\eqref{eq:traj} can be either an answer to the question or a temporal interval. We employ the same format in the multi-turn dialogue for two tasks. The unification enhances the model’s intrinsic grounding ability, which is fundamental to \emph{thinking-with-videos}. Accordingly, we adopt a mixture of VideoQA and grounding data for training. Please refer to Sec.~\ref{subsec:train-data} for more details.

\begin{figure}[!t]
    \centering
    \includegraphics[width=.95\linewidth]{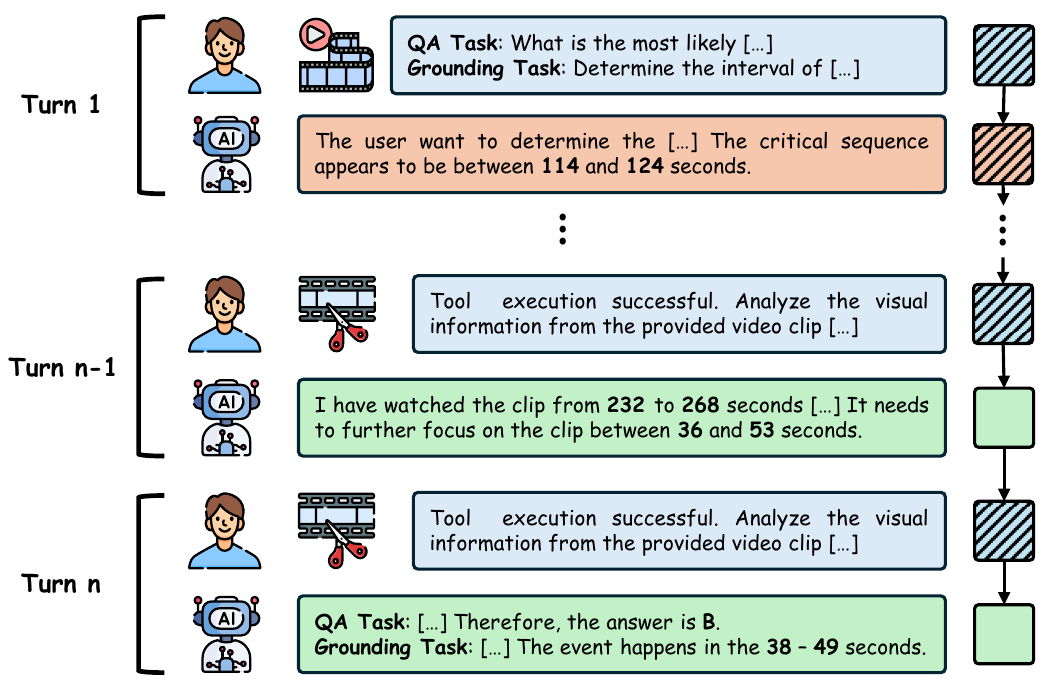}
    \caption{The unified masking mechanism, where only the last two turns of responses are supervised while others are masked.}
    \vspace{-10pt}
    \label{fig:unified_masking}
\end{figure}

\vspace{-10pt}
\paragraph{Unified Masking Strategy.}
In the collected tool-call data, the penultimate turn (Turn n-1 in Fig.~\ref{fig:unified_masking}) contains the correct temporal interval with the key evidence, and the final turn outputs the final answer. The temporal intervals in earlier turns are typically imprecise for multi-tool-call data, supervising these intervals during SFT can introduce noise.
Based on this insight, we apply the training loss only to the final two turns of the model outputs in the multi-turn dialogue, while all earlier generations and user inputs are masked so they do not affect gradients.

\begin{table*}[!t]
\setlength\tabcolsep{5pt}
\centering
\renewcommand{\arraystretch}{1.0}
\caption{Main comparison of long video understanding. Gemini-1.5-Pro and GPT-4o are included as closed-source model references. For fair comparisons, we ensure all methods use the same input resolution and maximum frames. $^\dagger$ denotes reproduced results using official checkpoints. \textbf{Bold} and \underline{underline} denotes the best and the second best results.}
\label{tab:main_long_video}
\resizebox{.9\linewidth}{!}{
\begin{tabular}{@{}lcccccccccc@{}}
\toprule
\multirow{3}{*}{Method} & MLVU & \multicolumn{4}{c}{VideoMMMU} & \multicolumn{4}{c}{VideoMME (w/o subtitle)} & LVBench \\
 & $\approx$ 651s & \multicolumn{4}{c}{$\approx$ 506s} & \multicolumn{4}{c}{$\approx$ 1018s} & $\approx$ 4101s \\ \cmidrule(l){2-2} \cmidrule(l){3-6} \cmidrule(l){7-10} \cmidrule(l){11-11} 
 & M-Avg & Adapt. & Compr. & Prec. & Avg. & Short & Medium & Long & Overall & Avg. \\ \midrule
\GR{10}Gemini-1.5-Pro~\cite{team2024gemini} & \multicolumn{1}{c|}{-} & 49.3 & 53.3 & 59.0 & \multicolumn{1}{c|}{53.8} & 81.7 & 74.3 & 67.4 & \multicolumn{1}{c|}{75.0} & 33.1 \\
\GR{10}GPT-4o~\cite{hurst2024gpt} & \multicolumn{1}{c|}{54.9} & 55.6 & 62.0 & 66.0 & \multicolumn{1}{c|}{61.2} & 80.0 & 70.3 & 65.3 & \multicolumn{1}{c|}{71.9} & 30.8 \\ \midrule
VILA-1.5-40B~\cite{lin2024vila} & \multicolumn{1}{c|}{44.2} & 32.6 & 30.6 & 38.6 & \multicolumn{1}{c|}{34.0} & - & - & - & \multicolumn{1}{c|}{60.1} & - \\
TimeMaker-8B~\cite{chen2024timemarker} & \multicolumn{1}{c|}{49.2} & - & - & - & \multicolumn{1}{c|}{-} & - & - & 46.4 & \multicolumn{1}{c|}{57.3} & {\ul 41.3} \\
VideoXL-7B~\cite{shu2025video} & \multicolumn{1}{c|}{45.5} & - & - & - & \multicolumn{1}{c|}{-} & - & - & - & \multicolumn{1}{c|}{55.5} & - \\
VideoRFT-7B~\cite{wang2025videorft} & \multicolumn{1}{c|}{-} & - & - & - & \multicolumn{1}{c|}{51.1} & - & - & - & \multicolumn{1}{c|}{59.8} & - \\
LongVA-7B~\cite{zhang2024long} & \multicolumn{1}{c|}{41.4} & - & - & - & \multicolumn{1}{c|}{-} & 61.1 & 50.4 & 46.2 & \multicolumn{1}{c|}{52.6} & - \\
LongVT-7B-RFT~\cite{yang2025longvt} & \multicolumn{1}{c|}{-} & 35.7 & 43.7 & 56.7 & \multicolumn{1}{c|}{45.3} & - & - & - & \multicolumn{1}{c|}{-} & {\ul 41.3} \\
Video-MTR-7B~\cite{xie2025video} & \multicolumn{1}{c|}{48.4} & - & - & - & \multicolumn{1}{c|}{-} & - & - & 51.0 & \multicolumn{1}{c|}{59.0} & - \\
VideoChat-R1-7B~\cite{li2025videochat} & \multicolumn{1}{c|}{45.2$^\dagger$} & 40.2$^\dagger$ & 45.6$^\dagger$ & 53.0$^\dagger$ & \multicolumn{1}{c|}{46.8$^\dagger$} & 72.2 & 60.7$^\dagger$ & 50.9$^\dagger$ & \multicolumn{1}{c|}{{\ul 62.1}} & 39.3$^\dagger$ \\
Video-R1-7B~\cite{feng2025video} & \multicolumn{1}{c|}{48.0$^\dagger$} & \textbf{44.2$^\dagger$} & 46.0$^\dagger$ & {\ul 67.3$^\dagger$} & \multicolumn{1}{c|}{{\ul 52.4}} & \textbf{74.1$^\dagger$} & {\ul 61.1$^\dagger$} & {\ul 51.2$^\dagger$} & \multicolumn{1}{c|}{61.4} & 40.1 \\
Qwen2.5-VL-7B~\cite{bai2025qwen2} & \multicolumn{1}{c|}{45.2$^\dagger$} & 35.9$^\dagger$ & 36.1$^\dagger$ & 57.6$^\dagger$ & \multicolumn{1}{c|}{43.2$^\dagger$} & 69.8$^\dagger$ & 59.2$^\dagger$ & 50.8$^\dagger$ & \multicolumn{1}{c|}{59.9$^\dagger$} & 39.2$^\dagger$ \\ \midrule
\RC{30}\Method-7B-SFT (Ours) & \multicolumn{1}{c|}{{\ul 49.5}} & 39.3 & \underline{46.4} & 60.4 & \multicolumn{1}{c|}{48.7} & 72.0 & 59.2 & 50.2 & \multicolumn{1}{c|}{60.6} & 39.6 \\
\RC{30}\Method-7B-RL (Ours) & \multicolumn{1}{c|}{\textbf{54.2}} & {\ul 43.0} & \textbf{47.8} & \textbf{69.0} & \multicolumn{1}{c|}{\textbf{53.2}} & {\ul 72.2} & \textbf{66.6} & \textbf{54.7} & \multicolumn{1}{c|}{\textbf{64.5}} & \textbf{43.0} \\ \bottomrule
\end{tabular}
}
\end{table*}

\subsection{Agentic Reinforcement Learning}
\label{subsec:method-rl}

\paragraph{Preliminary of GRPO.}
We adopt the on-policy variant of the GRPO algorithm~\cite{shao2024deepseekmath,guo2025deepseek}. For each training sample consisting of a video $V$ and a question $Q$, the current policy $\pi_\theta$ is used to sample $G$ multi-turn trajectories $\{\tau_1, \tau_2, \dots, \tau_G\}$.
The policy is optimized by maximizing the following objective:
\begin{equation}
\vspace{-5pt}
\begin{aligned}
    &\mathcal{J}_{GRPO}(\theta)= \mathbb{E}_{(V,Q)\sim D, \{\tau_i\}_{i=1}^G\sim\pi_\theta(\cdot\vert V,Q)} \\
    &\qquad\quad\frac{1}{\sum_{i=1}^G\vert \tau_i \vert}\sum_{i=1}^G\sum_{j=1}^{\vert\tau_i\vert}\big(A_{i,j}-\beta\mathbb{D}_{KL}[\pi_\theta\Vert\pi_{ref}]\big),
\end{aligned}
\end{equation}
where $\vert \tau_i \vert$ denotes the total number of tokens generated by the model in trajectory $i$, encompassing the thinking process, grounding outputs, and the final answer across all turns. After obtaining the final reward $\{r_i\}_{i=1}^G$ for each complete trajectory, the advantage term $A_{i,j}$ is normalized within the group and shared across all tokens $j$ within the same trajectory $\tau_i$: $A_{i,j}=\frac{r_i-mean(\{r_i\}_{i=1}^G)}{std(\{r_i\}_{i=1}^G)}$. The term $-\beta\mathbb{D}_{KL}[\pi_\theta\Vert\pi_{ref}]$ serves to constrain the deviation of the current policy $\pi_\theta$ from the reference policy $\pi_{ref}$. The reward $r_i$ measures the quality of the entire trajectory, including the effectiveness of grounding and reasoning content, as well as the correctness of the final answer.

\vspace{-12pt}
\paragraph{Reward Design.}
To jointly optimize answer correctness,  format adherence, and temporal grounding, we present a dedicated reward system consisting of three rewards.

\textbf{(1) Accuracy Reward.}
It measures answer correctness. The reward is 1 only if the response matches the ground truth.
\begin{equation}
\vspace{-5pt}
    R_\text{accuracy}=
\left\{
\begin{aligned}
1,\quad & \text{answer}=\text{gt},  \\
0,\quad & \text{answer}\ne \text{gt}.
\end{aligned}
\right.
\vspace{-5pt}
\end{equation}

\textbf{(2) Format Reward.}
This encourages the model to follow the required format throughout the multi-turn dialogue.
\begin{equation}
\vspace{-5pt}
R_\text{format}=
\left\{
\begin{aligned}
1,\quad & \text{format\ match},  \\
0,\quad & \text{otherwise}.
\end{aligned}
\right.
\vspace{-5pt}
\end{equation}

\textbf{(3) Penalty-aware IoU Reward.}
This is used to measure the quality of timestamp grounding. Let the model's predicted interval be $[t_s', t_e']$ and the ground truth interval be $[t_s, t_e]$, the grounding IoU is defined as:
\begin{equation}
R_\text{IoU}=
\frac{\vert [t_s,t_e] \vert\cap\vert [t_s', t_e'] \vert}{\vert [t_s,t_e] \vert\cup\vert [t_s', t_e'] \vert}.
\label{eq:reward-iou}
\end{equation}
When using IoU alone as a reward, the model may localize arbitrarily to hack the reward. To address this, we explicitly penalize these low-quality grounding behaviors. Specifically, when IoU is lower than a threshold $\sigma$, a penalty term $\lambda$ is applied to the response:
\begin{equation}
\label{eq:penalty_iou}
R_\text{penalty-IoU}=
\left\{
\begin{aligned}
&R_\text{IoU}-\lambda,\quad & R_\text{IoU} < \sigma,  \\
&R_\text{IoU},\quad & R_\text{IoU}\ge \sigma.
\end{aligned}
\right.
\end{equation}
This design effectively mitigates reward hacking and encourages the model to produce reliable groundings while discouraging arbitrary guesses, enabling more stable learning of multi-turn grounding and refinements.

\section{Experiment}
\label{sec:experiment}

\begin{table}[!t]
\setlength\tabcolsep{2pt}
\centering
\renewcommand{\arraystretch}{1.0}
\caption{Main comparison of temporal grounding. \textbf{Bold} and \underline{underline} denotes the best and the second best results.}
\vspace{-5pt}
\label{tab:main_temporal_grounding}
\resizebox{\linewidth}{!}{
\begin{tabular}{@{}lcccc@{}}
\toprule
\multirow{2}{*}{Method} & \multicolumn{2}{c}{Charades-STA} & \multicolumn{2}{c}{ActivityNet-MR} \\ \cmidrule(l){2-3} \cmidrule(l){4-5} 
 & R@0.7 & mIoU & R@0.7 & mIoU \\ \midrule
VTimeLLM-7B~\cite{huang2024vtimellm} & 11.4 & 31.2 & 14.3 & 30.4 \\
Momentor-7B~\cite{qian2024momentor} & 11.6 & 28.5 & 12.4 & 29.3 \\
ChatVTG-7B~\cite{qu2024chatvtg} & 15.8 & 34.8 & 9.4 & 27.2 \\
TimeMaker-8B~\cite{chen2024timemarker} & 26.9 & 48.4 & \textbf{33.0} & \textbf{49.5} \\
ZoomV-7B~\cite{pan2025timesearch} & 24.5 & 48.6 & 26.1 & 43.9 \\
VideoChat-R1-7B~\cite{li2025videochat}$^\dagger$ & \textbf{37.9} & {\ul 54.9} & 18.2 & 36.8 \\
Video-R1-7B~\cite{feng2025video}$^\dagger$ & 18.1 & 35.1 & 6.2 & 18.3 \\
Qwen2.5-VL-7B~\cite{bai2025qwen2}$^\dagger$ & 18.8 & 38.5 & 10.4 & 26.8 \\ \midrule
\RC{30}\Method-7B-SFT (Ours) & 28.0 & 48.9 & 24.7 & 41.1 \\
\RC{30}\Method-7B-RL (Ours) & {\ul 33.0} & \textbf{57.8} & {\ul 26.7} & {\ul 45.3} \\ \bottomrule
\end{tabular}
}
\end{table}

\subsection{Experimental Setup}

\textbf{Implementation Details.}
Following prior works, we choose Qwen2.5-VL-7B~\cite{bai2025qwen2} as the backbone. The training pipeline is built upon \textit{ms-swift}~\cite{zhao2024swiftascalablelightweightinfrastructure} and  \textit{vLLM}~\cite{kwon2023efficient}, with customized support for multi-round tool-call interactions. In Eq.~\ref{eq:penalty_iou}, we set the hyper-parameters $\lambda=0.1$ and $\sigma=0.1$. For additional training details, please refer to Appendix~\ref{appendix:training_config}.

\textbf{Benchmarks.}
We adopt a diverse set of public benchmarks for each task. For long video understanding, we evaluate on MLVU (test set)~\cite{zhou2025mlvu}, VideoMMMU~\cite{hu2025video}, VideoMME~\cite{fu2025video} and LVBench~\cite{wang2025lvbench}, which assess multimodal reasoning and comprehension over long video. For temporal grounding, we use Charades-STA~\cite{gao2017tall} and ActivityNet-MR~\cite{krishna2017dense}, which test the model's ability to localize temporal segments corresponding to text captions. For video GQA, we leverage NextGQA~\cite{xiao2024can} and ReXTime~\cite{chen2024rextime} to evaluate the ability to answer fine-grained, temporally grounded questions. Besides, we introduce VideoTemp-Bench to analyze performance across varying video lengths.

\subsection{Main Comparative Results}

\textbf{\Method achieves remarkable long video understanding.}
As shown in Tab.~\ref{tab:main_long_video}, \Method achieves state-of-the-art performance across nearly all long video understanding benchmarks. For example, our method obtains 2.4\% and 1.7\% improvements on VideoMME and LVBench, respectively, indicating strong multimodal understanding in long videos. The initial SFT stage incentivizes fundamental patterns for basic video understanding using curated trajectories, and subsequent RL consistently enhances video reasoning, \ie, by roughly 4$\sim$5\% across all benchmarks.

\textbf{\Method exhibits strong internal temporal grounding abilities.}
\Method also delivers strong temporal grounding performance, even comparable to that of expert grounding models like TimeMaker~\cite{chen2024timemarker}, as shown in Tab.~\ref{tab:main_temporal_grounding}. The reliable grounding is a prerequisite for the implementation of \emph{thinking-with-videos}.

\textbf{The \emph{thinking-with-videos} paradigm ensures strong GQA performance.}
As indicated in Tab.~\ref{tab:main_gqa}, \Method attains top-tier performance on both NextGQA and ReXTime. Notably, the model consistently attains higher mIoU and answer accuracy, indicating more precise evidence grounding and highlighting how improved localization contributes to stronger video understanding.
The consistent gains validate the effectiveness of the on-demand tool-call abilities in \emph{thinking-with-videos}. By identifying and focusing on the most relevant video segments, the model significantly enhances long video perception and reasoning.

\begin{table}[!t]
\setlength\tabcolsep{2pt}
\centering
\renewcommand{\arraystretch}{1.0}
\caption{Main comparison of video GQA. \textbf{Bold} and \underline{underline} denotes the best and the second best results.}
\label{tab:main_gqa}
\resizebox{\linewidth}{!}{
\begin{tabular}{@{}lcccc@{}}
\toprule
\multirow{2}{*}{Method} & \multicolumn{2}{c}{NextGQA} & \multicolumn{2}{c}{ReXTime} \\ \cmidrule(l){2-3} \cmidrule(l){4-5} 
 & mIoU & Acc & mIoU & Acc \\ \midrule
VTimeLLM-7B~\cite{huang2024vtimellm} & - & - & 20.1 & 36.1 \\
TimeChat-7B~\cite{ren2024timechat} & - & - & 11.6 & 40.0 \\
DeepVideo-R1-7B~\cite{park2025deepvideo} & \textbf{36.8} & 72.5 & \textbf{-} & \textbf{-} \\
VideoChat-R1-7B~\cite{li2025videochat}$^\dagger$ & 32.4 & 70.6 & 10.3 & 71.1 \\
Video-R1-7B~\cite{feng2025video}$^\dagger$ & 17.5 & 74.3 & 7.3 & 70.5 \\
Qwen2.5-VL-7B~\cite{bai2025qwen2}$^\dagger$ & 22.7 & 74.8 & 8.7 & {\ul 73.4} \\ \midrule
\RC{30}\Method-7B-SFT (Ours) & 30.3 & {\ul 75.4} & {\ul 23.7} & 71.4 \\
\RC{30}\Method-7B-RL (Ours) & {\ul 33.4} & \textbf{76.4} & \textbf{29.5} & \textbf{74.4} \\ \bottomrule
\end{tabular}
}
\end{table}

\begin{table}[!t]
\setlength\tabcolsep{2pt}
\centering
\renewcommand{\arraystretch}{1.0}
\caption{Ablations on \colorbox{color2}{training components in the SFT stage} and \colorbox{color3}{rewards in the RL stage} in comparison with \colorbox{color1}{our full model}.}
\label{tab:ablation-sft-rl}
\resizebox{\linewidth}{!}{
\begin{tabular}{@{}llccccc@{}}
\toprule
\multirow{2}{*}{ID} & \multirow{2}{*}{Variants} & VideoMMMU & VideoMME & LVBench & \multicolumn{2}{c}{ReXTime} \\ \cmidrule(l){3-7} 
 &  & Overall & Overall & Overall & mIoU & Acc \\ \midrule
\cellcolor{color1}(a) & \cellcolor{color1}\Method & \textbf{53.2} & \textbf{64.5} & \textbf{43.0} & \textbf{29.5} & \textbf{74.4} \\ \midrule
\cellcolor{color2}(b) & \cellcolor{color2}w/o grounding data & \underline{52.5} & 63.0 & \underline{42.0} & 13.0 & 73.3 \\
\cellcolor{color2}(c) & \cellcolor{color2}w/o unified masking & 47.9 & 61.5 & 41.2 & 18.8 & 70.6 \\ \midrule
\cellcolor{color3}(d) & \cellcolor{color3}w/o IoU reward & 51.6 & 63.3 & 41.7 & \underline{26.2} & \underline{73.7} \\
\cellcolor{color3}(e) & \cellcolor{color3}w/o penalty-aware & 44.2 & \underline{63.7} & 40.7 & 23.8 & 73.6 \\ \bottomrule
\end{tabular}
}
\end{table}

\begin{table*}[!t]
\setlength\tabcolsep{6pt}
\centering
\renewcommand{\arraystretch}{1.0}
\caption{Performance of temporal grounding (mIoU) and VideoQA (Acc) of various durations in VideoTemp-Bench.}
\label{tab:videotemp_bm}
\resizebox{.9\linewidth}{!}{
\begin{tabular}{@{}lcccccccccc@{}}
\toprule
\multirow{2}{*}{Method} & \multicolumn{2}{c}{0$\sim$3min} & \multicolumn{2}{c}{3$\sim$10min} & \multicolumn{2}{c}{10$\sim$20min} & \multicolumn{2}{c}{$>$ 20min} & \multicolumn{2}{c}{Overall} \\ \cmidrule(l){2-3} \cmidrule(l){4-5} \cmidrule(l){6-7} \cmidrule(l){8-9} \cmidrule(l){10-11} 
 & mIoU & Acc & mIoU & Acc & mIoU & Acc & mIoU & Acc & mIoU & Acc \\ \midrule
\GR{10}Gemini-2.5-Pro~\cite{comanici2025gemini} & 39.1 & 81.6 & 46.1 & 91.3 & 36.1 & 90.0 & 14.8 & 76.1 & 34.0 & 84.7 \\ \midrule
VideoChat-R1-7B~\cite{li2025videochat} & 25.2 & \textbf{82.0} & 6.7 & \underline{87.6} & 4.7 & 73.6 & 1.8 & \underline{53.6} & 9.6 & \underline{74.2} \\
Video-R1-7B~\cite{feng2025video} & 13.3 & \underline{81.0} & 4.0 & 86.0 & 4.0 & \underline{75.3} & 2.1 & 53.3 & 5.9 & 73.9 \\
Qwen2.5-VL-7B~\cite{bai2025qwen2} & 17.6 & 75.0 & 9.1 & 81.6 & 6.9 & 70.3 & 1.0 & 49.1 & 8.7 & 69.1 \\ \midrule
\RC{30}\Method-7B-SFT (Ours) & \underline{33.7} & 78.3 & \underline{18.6} & 79.5 & \underline{14.6} & 68.5 & \underline{3.3} & \textbf{54.6} & \underline{17.8} & 70.5 \\
\RC{30}\Method-7B-RL (Ours) & \textbf{35.3} & \underline{81.0} & \textbf{32.0} & \textbf{90.6} & \textbf{21.7} & \textbf{81.3} & \textbf{4.0} & \textbf{54.6} & \textbf{23.4} & \textbf{77.0} \\ \bottomrule
\end{tabular}
}
\end{table*}

\subsection{Ablation Studies}

\textbf{Ablations on grounding training data.}
As Tab.~\ref{tab:ablation-sft-rl} shows, SFT training with grounding data not only directly enhances grounding performance but also implicitly improves video QA accuracy. For example, (a) outperforms (b) 16.5\% and 1.1\% on mIoU and Acc of ReXTime, respectively.

\textbf{Ablations on unified masking mechanism.}
Removing the unified masking strategy and instead supervising all responses during SFT leads to significant performance degradation, \ie, (a) $\to$ (c) in Tab.~\ref{tab:ablation-sft-rl}. We attribute this degradation to the noise introduced by unmasked, incorrect reasoning paths of the initial coarse grounding turn. These results underscore the importance of our unified masking strategy in stabilizing training using correct signals.

\begin{figure}[!t]
  \centering
  \begin{subfigure}{0.435\columnwidth}
    \centering
    \includegraphics[width=\linewidth]{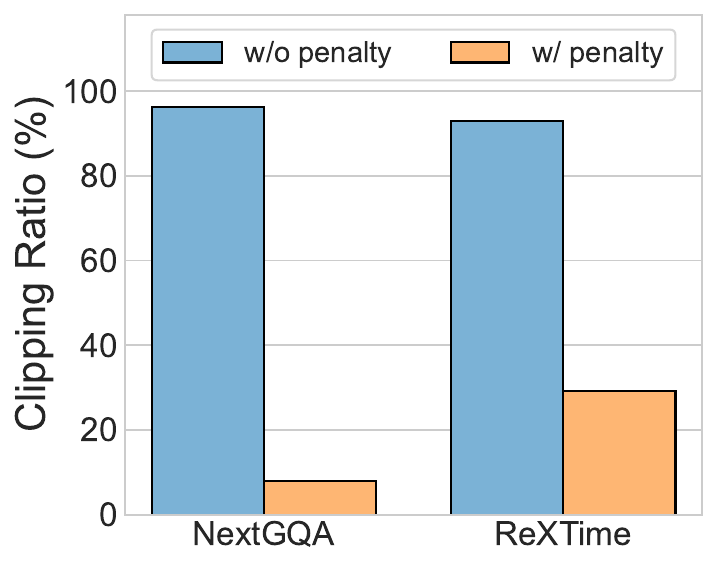}
    \caption{Video clipping ratio.}
    \label{fig:hack-ratio}
  \end{subfigure}
  \hspace{0.02\columnwidth}
  \begin{subfigure}{0.435\columnwidth}
    \centering
    \includegraphics[width=\linewidth]{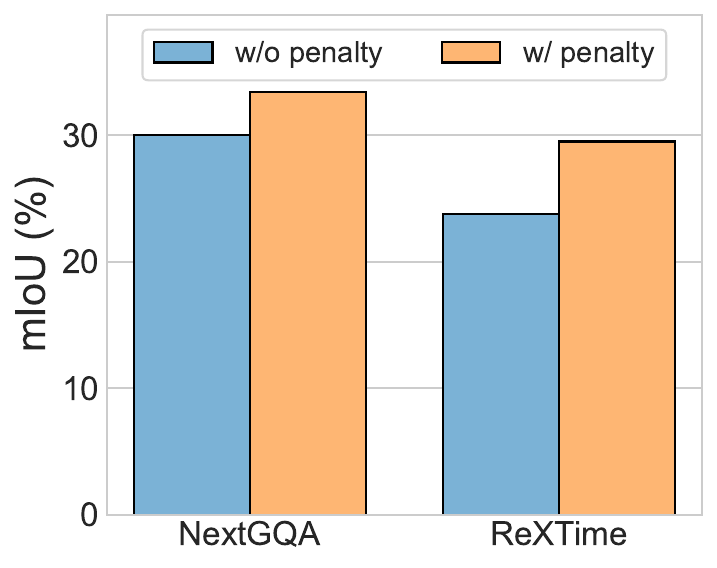}
    \caption{mIoU.}
    \label{fig:hack-iou}
  \end{subfigure}
  \caption{Reward hacking with native IoU rewards.}
  \label{fig:reward-hacking}
  \vspace{-10pt}
\end{figure}

\textbf{Naive IoU reward can readily induce reward hacking.}
Replacing the penalty-aware IoU reward $R_\text{penalty-IoU}$ with a standard IoU reward $R_\text{IoU}$ in Eq.~\eqref{eq:reward-iou} leads to performance degradation in Tab.~\ref{tab:ablation-sft-rl}. We delve into this phenomenon in Fig.~\ref{fig:reward-hacking}. $R_{\text{IoU}}$ causes the tool-call ratio to rise sharply (Fig.~\ref{fig:hack-ratio}) while substantially degrading grounding quality (Fig.~\ref{fig:hack-iou}). In short, $R_\text{IoU}$ without any penalty encourages the model to localize blindly to hack IoU rewards while neglecting grounding performance.

\textbf{Ablations on IoU reward.}
In the RL stage, removing all IoU rewards (only keeping $R_\text{acc}$ and $R_\text{format}$) in Tab.~\ref{tab:ablation-sft-rl} (d) yields a modest decline in both tasks, \eg, a 1.6\% decline in VideoMMMU. The effect of the IoU reward is similar to the grounding data in SFT, \ie, enhancing the model's internal grounding abilities for desirable \emph{thinking-with-video} behaviors.

\subsection{Further Analysis}

\begin{figure}[!t]
  \centering
  \begin{subfigure}{0.48\columnwidth}
    \centering
    \includegraphics[width=\linewidth]{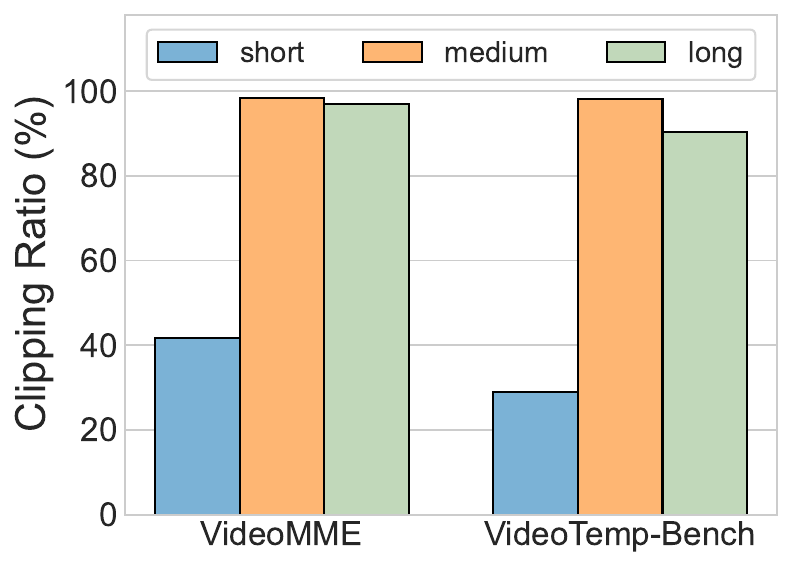}
    \caption{Clipping ratio.}
    \label{fig:on-demand-tool-ratio}
  \end{subfigure}
  \hspace{0.02\columnwidth}
  \begin{subfigure}{0.48\columnwidth}
    \centering
    \includegraphics[width=\linewidth]{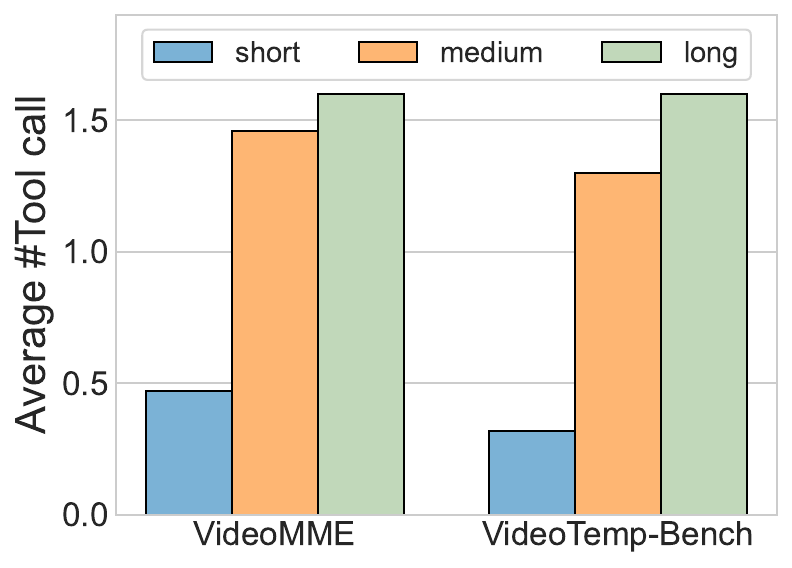}
    \caption{Average \#clip per video.}
    \label{fig:on-demand-tool-num}
  \end{subfigure}
  \caption{On-demand tool-call according to video length.}
  \label{fig:on-demand-tool}
  \vspace{-10pt}
\end{figure}

\textbf{Our method could perform on-demand tool-call.}
We explicitly measure the tool-call statistics of \Method across various video durations. Fig.~\ref{fig:on-demand-tool-ratio} shows the tool-call rate for medium and long videos is significantly higher than that for short videos. This trend is accompanied by a corresponding increase in the average number of cropping operations (Fig.~\ref{fig:on-demand-tool-num}). These results collectively indicate that, for longer videos, where informative cues are sparser, our model is more inclined to invoke video clipping to localize relevant visual evidence. In this regard, our method is able to perform on-demand clipping flexibly.

\begin{figure*}[!t]
    \centering
    \includegraphics[width=\linewidth]{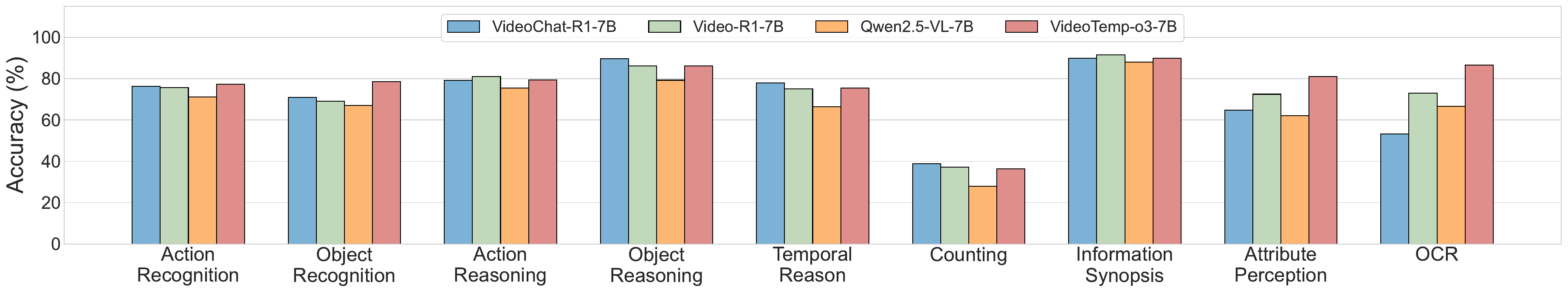}
    \caption{Performance of different video tasks in VideoTemp-Bench.}
    \label{fig:videotemp-bench-task}
    \vspace{-5pt}
\end{figure*}

\textbf{Temporal grounding remains a key bottleneck for long video understanding.}
In Tab.~\ref{tab:videotemp_bm}, we evaluate several models on our VideoTemp-Bench containing videos of various durations. Generally, on longer videos (\eg, $>$20min) where visual evidence is sparser, grounding performance drops remarkably, and QA accuracy deteriorates accordingly. For example, when video duration increases from $<$10min to $>$20min, the performance Qwen2.5-VL drops from 70.3\% to 49.1\%. We hypothesize that this degradation mainly comes from the sparse initial sampling required by context-length and computational constraints: if the key evidence is not observed during the initial skim, the model may fail to identify a reliable region for later refinement. In addition, hour-long videos often contain redundant scenes and transitions that further distract temporal localization. Our \emph{thinking-with-videos} paradigm alleviates this issue to some extent. Nevertheless, accurate temporal localization remains a critical challenge for long videos. To further investigate this issue, we conducted additional experiments analyzing the effect of using ground truth video segments. Details can be found in Appendix~\ref{appendix:impact_of_temporal_grounding}.

\textbf{Beyond grounding, certain types of tasks remain a major weakness for video understanding.}
We comprehensively evaluate video understanding across a wide range of tasks in Fig.~\ref{fig:videotemp-bench-task}. The results show substantial performance variability across different video tasks. Generally, complex reasoning tasks (temporal reasoning) and fine-grained perception (counting and OCR) are more challenging, whereas basic perception tasks (information synopsis and recognition) are comparatively easier. Specifically, there is an approximately 40\% gap between counting and recognition tasks.

\textbf{Learned tool-use behavior is more effective than fixed or prompt-only strategies.}
To further isolate the effect of our agentic inference process, we compare \Method with several inference-time controls in Tab.~\ref{tab:inference_controls}. The first two control strategies are applied to Qwen2.5-VL-7B, which does not natively include a localize-clip-answer process. Prompt-only re-grounding and inference-time self-correction show only marginal differences from direct answering, indicating that prompt-level interventions alone provide limited gains. The latter two controls are applied to \Method: always cropping once and directly answering both underperform the full model, especially on medium and long videos where relevant evidence is more sparse. These results indicate that explicit trajectory training and reward optimization are important for learning when to crop, when to refine, and when to answer.

\begin{table}[!t]
\setlength\tabcolsep{3pt}
\centering
\renewcommand{\arraystretch}{1.0}
\caption{Inference-time control comparisons on VideoMME (w/o subtitle). The first two controls are applied to Qwen2.5-VL-7B, while the next two controls are applied to \Method.}
\label{tab:inference_controls}
\resizebox{\linewidth}{!}{
\begin{tabular}{@{}lcccc@{}}
\toprule
Method & Short & Medium & Long & Overall \\ \midrule
Prompt-only re-grounding & 69.0 & 56.3 & 49.8 & 58.4 \\
Inference-time self-correction & 64.7 & 63.4 & \underline{53.2} & 60.5 \\
Qwen2.5-VL-7B & 69.8 & 59.2 & 50.8 & 59.9 \\
\midrule
w/ crop once & \underline{72.0} & \underline{64.0} & 53.1 & \underline{63.0} \\
w/o tool call & 70.3 & 63.4 & 52.7 & 62.1 \\
\Method & \textbf{72.2} & \textbf{66.6} & \textbf{54.7} & \textbf{64.5} \\ \bottomrule
\end{tabular}
}
\end{table}
\section{Conclusion}
\label{sec:conclusion}

In this work, we study the agentic \emph{thinking-with-videos} paradigm for long video understanding and propose a unified framework, namely \Method. It follows the pipeline of \emph{localize-clip-answer}, by proactively cropping query-relevant video segments, the model could produce reliable answers grounded in key visual features.
\Method is characterized by three features: on-demand video cropping, a reflection mechanism to refine imprecise grounding, and the unification of both temporal grounding and VideoQA tasks. To achieve this, we propose a unified masking mechanism in the SFT stage to incentivize desirable behaviors and devise dedicated rewards to mitigate reward hacking. Besides, we introduce a data curation pipeline to curate high-quality multi-turn trajectories for SFT and reliable long video GQA annotation for RL.
Owing to these components, our method outperforms previous methods across a broad spectrum of long video QA and grounding benchmarks. 
Future work could incorporate a broader suite of external tools, such as search engines and spatiotemporal grounding operations, to enable the model to tackle complex and up-to-date tasks in pragmatic scenarios.

\section{Limitations}
\label{sec:limitations}

Although \Method improves long-video understanding through iterative temporal grounding, several limitations remain. For very long videos, the initial skim is still sparse due to context-length and computational constraints, so key evidence may be missed before refinement. This issue is more severe for extremely short events that last only a few frames. In addition, the localize-clip-answer process introduces sequential inference overhead, and the data construction pipeline still benefits from strong teacher models and verified temporal annotations, which may increase construction cost. Our unified masking strategy also assumes that later verified turns are more reliable than earlier exploratory turns; if future data construction can provide accurate supervision for every turn, this heuristic may need to be adapted. Future work could explore hierarchical sampling, finer temporal modeling, parallel clip verification, and weaker forms of grounding supervision.

\section*{Acknowledgements}
\label{sec:acknowledgements}

This work was supported in part by the National Natural Science Foundation of China (Nos. 62376140, 62376137, U23A20315, and 62572282), the Shandong Provincial Natural Science Foundation (No. ZR2022YQ59), the Science and Technology Innovation Program for Distinguished Young Scholars of Shandong Province Higher Education Institutions (No. 2023KJ128), the Special Fund for Taishan Scholar Project of Shandong Province, the Key R\&D Program of Shandong Province, China (No. 2025CXGC020101), and Kuaishou (No. DJHL-20240801-004), providing essential funding and valuable institutional support to this research.


\section*{Impact Statement}
\label{sec:impact_statements}


This paper aims to advance long-video understanding by improving temporal grounding and evidence-based video question answering. Such techniques may support beneficial applications such as video retrieval, educational content analysis, accessibility tools, and assistance for reviewing long-form visual records. At the same time, more capable video understanding systems may also be misused for intrusive surveillance, privacy-sensitive profiling, or automated decision making without appropriate human oversight.

Our method can still produce incorrect temporal localizations or answers, especially for very long videos, ambiguous questions, or short-duration events. Therefore, it should not be directly deployed in high-stakes scenarios where incorrect video interpretation may cause harm. In addition, the data construction process relies on strong teacher models and human verification, and may inherit biases from source datasets or teacher annotations. We encourage future work to study robustness, privacy protection, bias mitigation, and human-in-the-loop verification when applying long-video understanding models in real-world settings.

\nocite{langley00}

\bibliography{main}
\bibliographystyle{icml2026}


\newpage
\appendix
\onecolumn



\section{Further Details about Dataset and Benchmark}
\label{appendix:dataset_and_benchmark}
\begin{figure}[!ht]
    \centering
    \begin{subfigure}[t]{.24\linewidth}
        \centering
        \includegraphics[width=\linewidth]{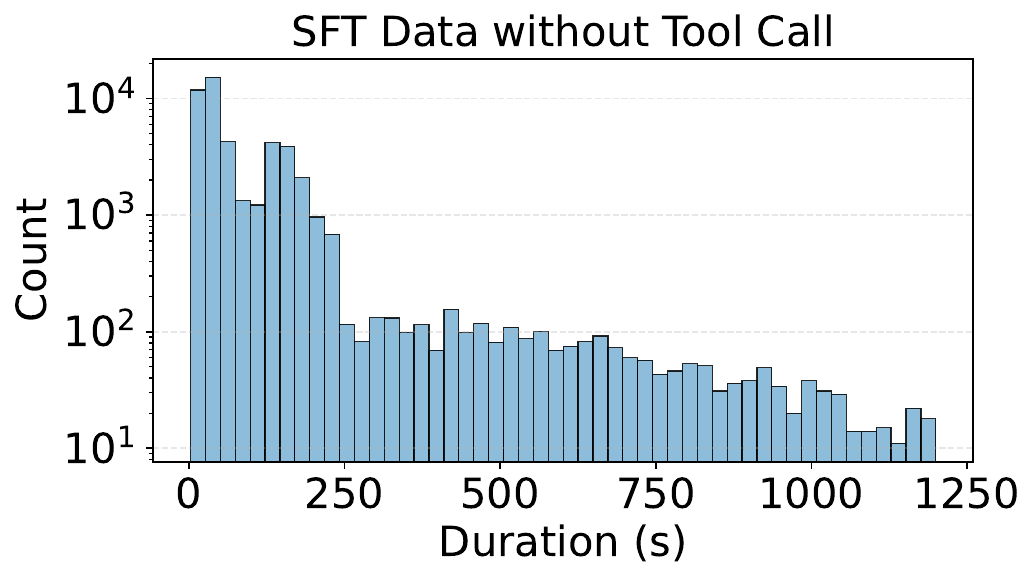}
        \caption{SFT data without tool call.}
        \label{fig:duration_sft_wo_tool}
    \end{subfigure}\hfill
    \begin{subfigure}[t]{.24\linewidth}
        \centering
        \includegraphics[width=\linewidth]{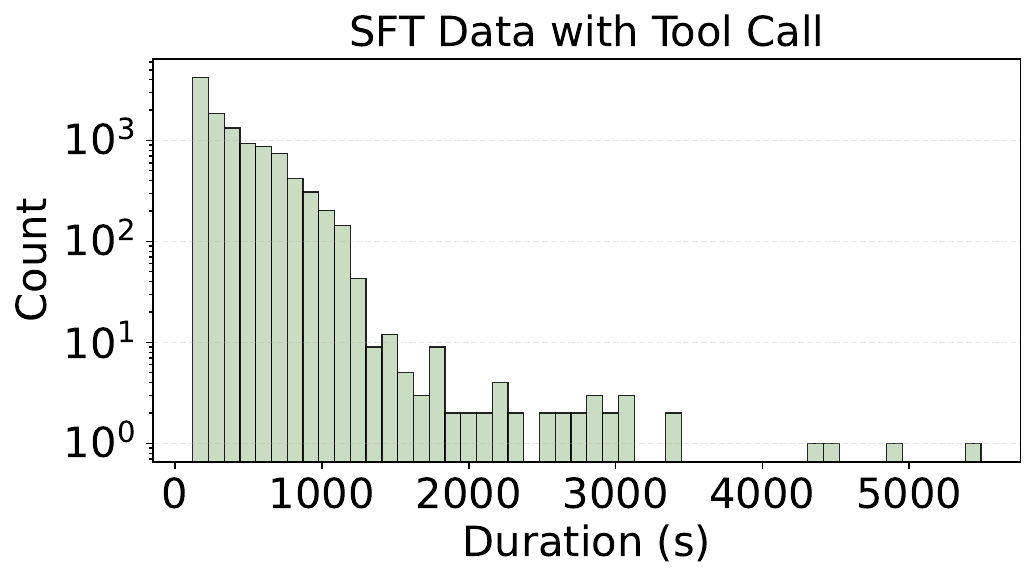}
        \caption{SFT data with tool call.}
        \label{fig:duration_sft_wi_tool}
    \end{subfigure}\hfill
    \begin{subfigure}[t]{.24\linewidth}
        \centering
        \includegraphics[width=\linewidth]{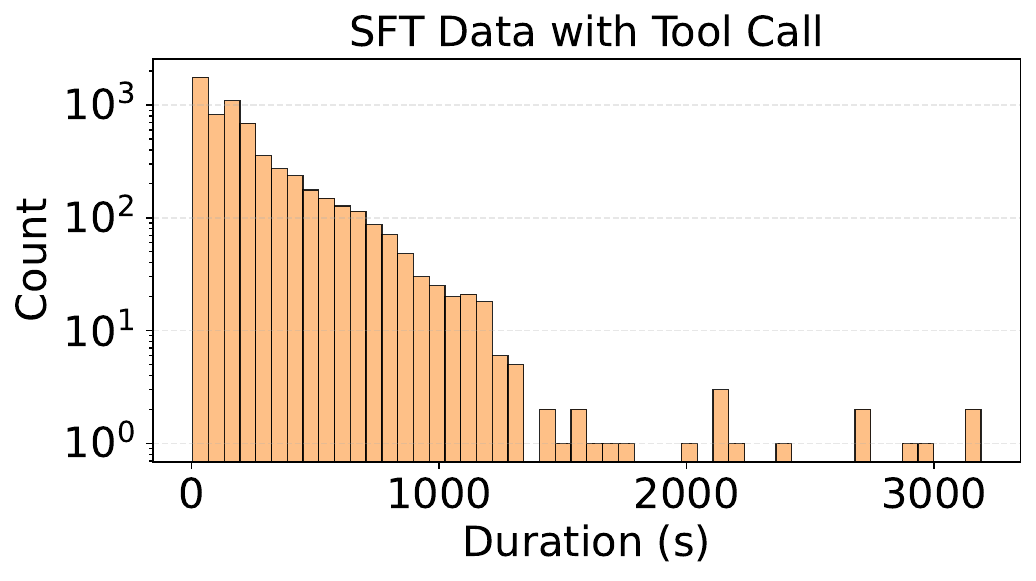}
        \caption{RL data.}
        \label{fig:duration_rl}
    \end{subfigure}\hfill
    \begin{subfigure}[t]{.24\linewidth}
        \centering
        \includegraphics[width=\linewidth]{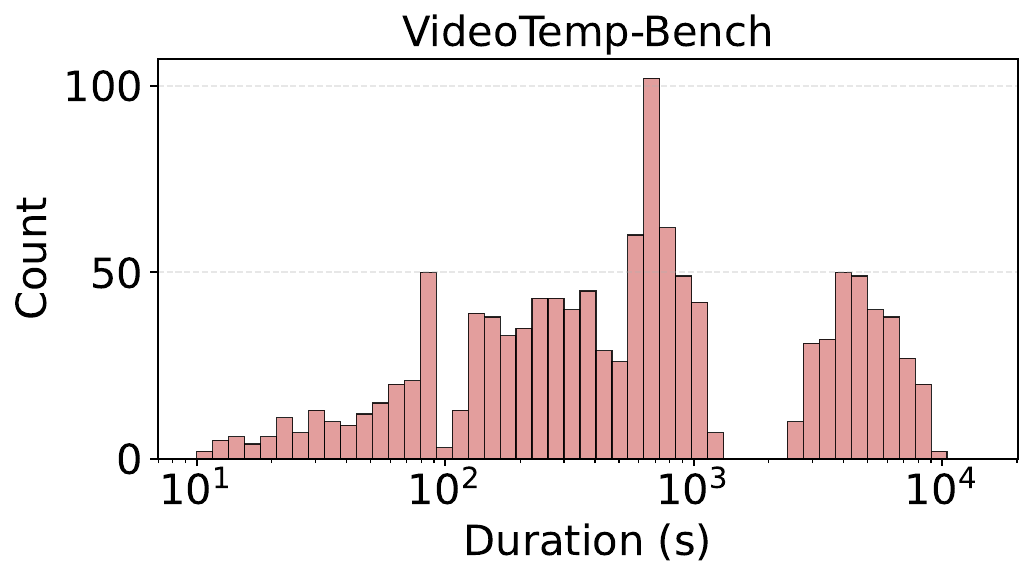}
        \caption{VideoTemp-Bench.}
        \label{fig:duration_videotemp}
    \end{subfigure}

    \caption{Duration distribution of training data and benchmark.}
    \label{fig:duration}
\end{figure}

To provide a more comprehensive understanding of the training data and VideoTemp-Bench, we present the duration distributions of various data subsets in Fig.~\ref{fig:duration}. 

As shown in Fig.~\ref{fig:duration_sft_wo_tool} and Fig.~\ref{fig:duration_sft_wi_tool}, the durations of SFT data without tool calls are concentrated in shorter video ranges. This aligns with the intuition that short videos often contain sufficient visual context, making explicit localization unnecessary. In contrast, longer videos require the model to process and filter more complex visual content, which motivates the construction of multi-turn tool-call samples specifically for such cases. For the RL phase, as shown in Fig.~\ref{fig:duration_rl}, we adopt a diverse distribution of video durations. This broader coverage ensures the model is exposed to various temporal complexities and better equipped to learn on-demand tool invocation strategies in different video comprehension scenarios.

\begin{figure}[!ht]
    \centering
    \includegraphics[width=0.9\linewidth]{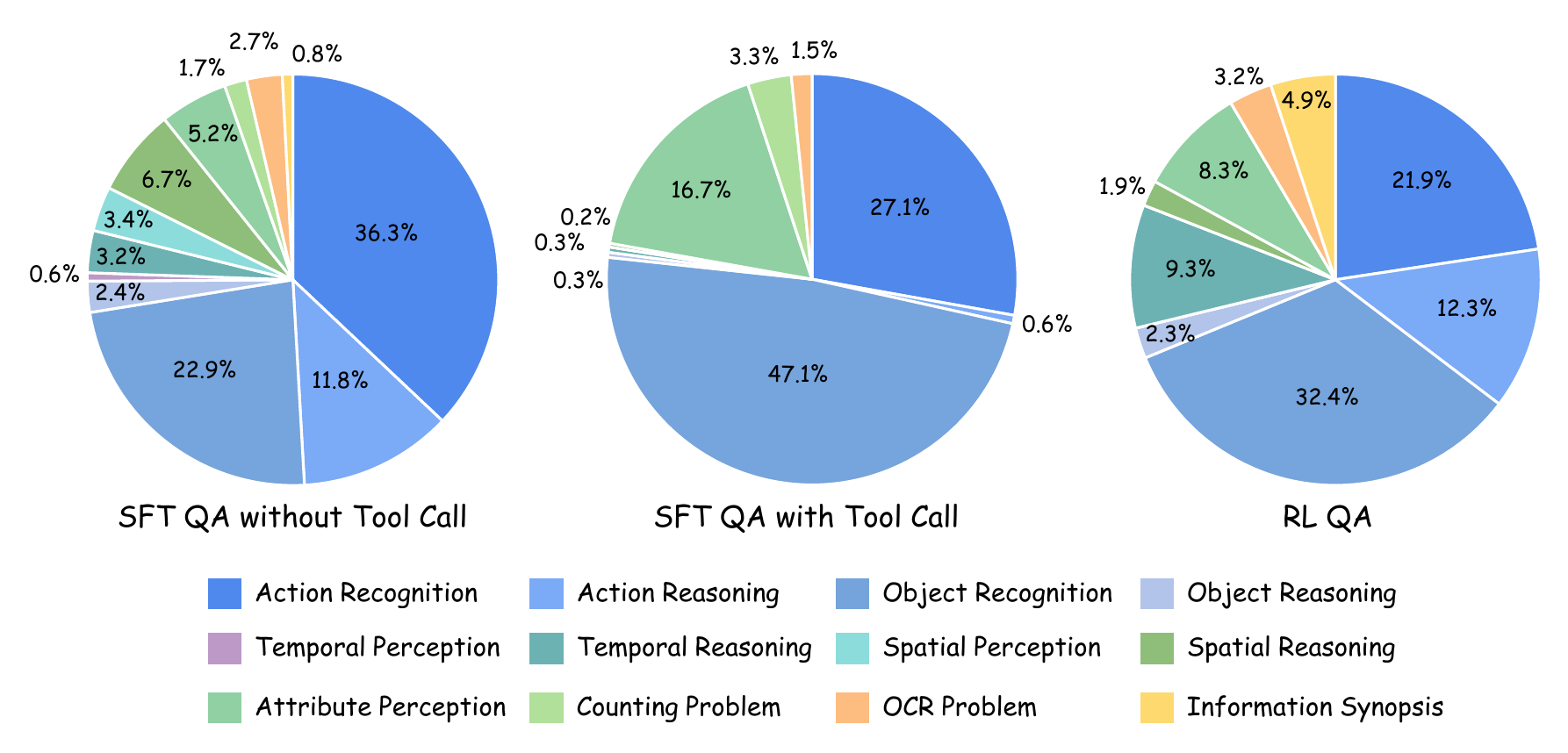}
    \caption{Distribution of question type in QA training data.}
    \label{fig:qa_category}
\end{figure}

Beyond video duration distribution, we also present the question type distribution across QA datasets, as illustrated in Fig.~\ref{fig:qa_category}. For the SFT QA data without tool calls, we incorporate a diverse range of question types to enhance the model's general comprehension and reasoning capabilities. In contrast, the SFT QA data with tool calls primarily focuses on question types such as Action/Object Recognition and Attribution Perception, which heavily rely on grounding within specific video segments. This targeted design aims to better train and strengthen the model's ability to perform \emph{thinking-with-videos} reasoning. In the RL data, while maintaining a high proportion of the aforementioned grounding-focused categories, we further introduce a broader variety of question types. This diversity encourages the model to autonomously explore, adopt, and apply the \emph{thinking-with-videos} paradigm more effectively throughout the RL process.

Regarding VideoTemp-Bench, as mentioned in Sec.~\ref{subsec:videotemp-bench}, the benchmark is composed of four video length categories: 0$\sim$3 minutes, 3$\sim$10 minutes, 10$\sim$20 minutes, and over 20 minutes. The duration distribution for this benchmark is detailed in Fig.~\ref{fig:duration_videotemp}. Specifically, samples in the 0$\sim$3 minute range are drawn from NextGQA and LongVILA; the 3$\sim$10 minute and 10$\sim$20 minute intervals utilize data from LongVILA and LongVideo-Reason~\cite{chen2025scaling}; and videos longer than 20 minutes are sourced from ScaleLong~\cite{ma2025scalelong}. Each duration group contains 300 samples, totaling 1,200 samples. This stratified sampling design ensures both broad coverage and comparability, enabling stable and interpretable evaluation of model performance across varying video lengths. Compared with CG-Bench~\cite{chen2025cgbench}, which focuses on clue-grounded QA for long videos, VideoTemp-Bench explicitly evaluates both temporal localization and answer accuracy across a wider range of video durations.

\begin{table}[!t]
\centering
\setlength\tabcolsep{3pt}

\begin{subtable}[t]{0.48\linewidth}
  \centering
  \caption{Training configurations.}
  \label{tab:training_config}
  \scalebox{0.88}{
    \begin{tabular}{@{}lcc@{}}
    \toprule
    Configuration & SFT & RL \\ \midrule
    Learning rate & 1e-5 & 5e-6 \\
    Batch size & 256 & 72 \\
    Epochs & 3 & 1 \\
    Group size & - & 8 \\
    Max turns & 3 & 3 \\
    Max frames & 512 & 512 \\
    Max pixels & 224$\times$224 & 224$\times$224 \\
    Tool call enabled & Yes & Yes \\
    Training mode & full & full \\
    \bottomrule
    \end{tabular}
  }
\end{subtable}\hfill
\begin{subtable}[t]{0.48\linewidth}
  \centering
  \caption{Evaluation configurations for each benchmark.}
  \label{tab:eval_config}
  \scalebox{0.88}{
    \begin{tabular}{@{}lcc@{}}
    \toprule
    Benchmark & Max Frames & Max Pixels \\ \midrule
    MLVU~\cite{zhou2025mlvu} & 256 & 448$\times$448 \\
    VideoMMMU~\cite{hu2025video} & 256 & 448$\times$448 \\
    VideoMME~\cite{fu2025video} & 1024 & 224$\times$224 \\
    LVBench~\cite{wang2025lvbench} & 1024 & 224$\times$224 \\
    Charades-STA~\cite{gao2017tall} & 256 & 448$\times$448 \\
    ActivityNet-MR~\cite{krishna2017dense} & 1024 & 224$\times$224 \\ 
    NextGQA~\cite{xiao2024can} & 256 & 448$\times$448 \\
    ReXTime~\cite{chen2024rextime} & 1024 & 224$\times$224 \\
    VideoTemp-Bench (Ours) & 1024 & 224$\times$224 \\
    \bottomrule
    \end{tabular}
  }
\end{subtable}

\vspace{7pt}

\caption{Configurations used in training and evaluation.}
\label{tab:configs}
\end{table}

\section{Experimental Details}
\subsection{Training Details}
\label{appendix:training_config}
For SFT and RL, we set the learning rates to 1e-5 and 5e-6, respectively. All videos are sampled at a maximum rate of 2 FPS, with each original video uniformly sampled to up to 512 frames, and each clipped video sampled to a maximum of 64 frames. Each frame is constrained to a maximum resolution of 224$\times$224 pixels. We use a batch size of 256 for SFT. For RL, each training step involves 72 samples, with each sample consisting of a group of 8 trajectories using GRPO. All the training is conducted on eight GPUs. For more details, please refer to Tab.~\ref{tab:training_config}.

\subsection{Evaluation Details}
\label{appendix:eval_config}
During benchmark evaluation, videos are sampled at a maximum of 2 FPS, and the generation temperature is set to 0.1. Each dialogue session supports up to 3 conversational turns, allowing for one round of timestamp refinement. The detailed evaluation configurations for each benchmark are summarized in Tab.~\ref{tab:eval_config}.

\begin{figure}[!t]
    \centering
    \includegraphics[width=0.9\linewidth]{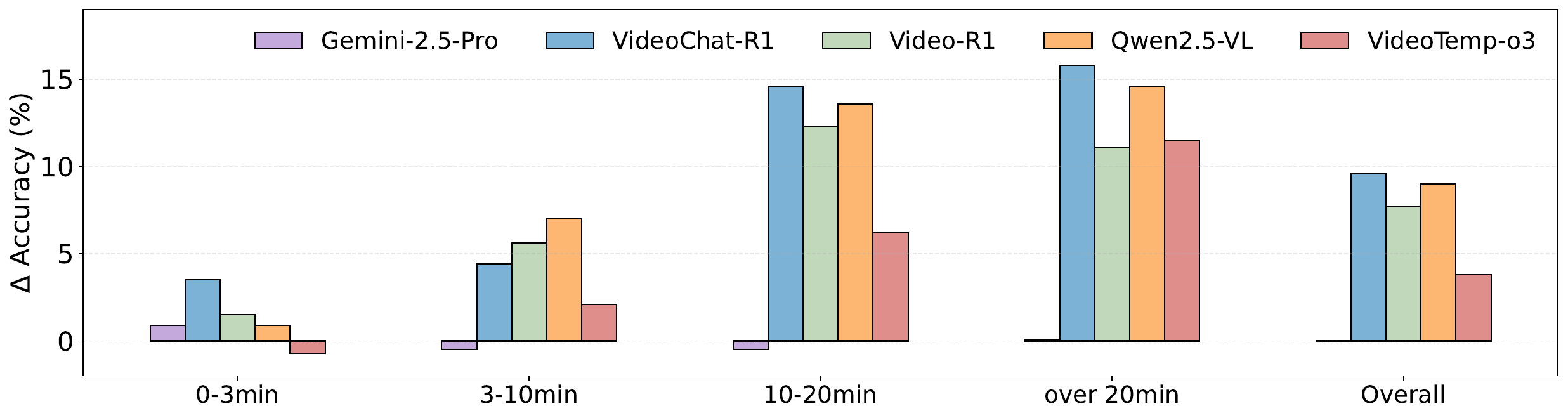}
    \caption{Accuracy change across video lengths when using ground truth video segments.}
    \label{fig:delta}
    \vspace{-10pt}
\end{figure}

\section{Impact of temporal grounding on VideoQA}
\label{appendix:impact_of_temporal_grounding}
To better understand the role of temporal grounding in VideoQA, we conduct an analysis by replacing the original videos in VideoTemp-Bench with the ground truth video segments corresponding to the annotated timestamps. This allows us to isolate and evaluate the effect of precise temporal grounding on model performance.

As shown in Fig.~\ref{fig:delta}, Qwen2.5-VL, VideoChat-R1, and Video-R1 exhibits a consistent performance improvement across all video length categories when evaluated using the ground truth video segments. The performance gains become more pronounced with increased video length, suggesting that longer videos pose greater challenges for segment localization. Consequently, access to accurately cropped content plays a more critical role in enabling precise reasoning and answer generation in such cases.

In comparison, \Method also achieves performance improvements under ground truth localization. However, the magnitude of this improvement is much smaller. Particularly for short videos (0-3 minutes), the gains are minimal. This can be attributed to \Method's internalization of the \emph{thinking-with-videos} paradigm, which enables it to effectively identify and reason over relevant temporal segments during inference. As a result, its dependence on externally provided localization is reduced, especially in shorter videos where the temporal space is inherently limited.

In contrast, Gemini-2.5-Pro shows negligible performance difference across all video length. This is likely due to its large context window and strong video understanding capabilities. Even with hour-long video inputs, Gemini-2.5-Pro can attend to nearly the entire sequence and accurately localize key information without the need for explicit cropping. This highlights the importance of dense frame sampling in long-form video understanding, which allows powerful models to maintain high reasoning accuracy across an extended temporal context.

\section{Reward Dynamics During RL Training.}

To more precisely evaluate the significance of the penalty-aware IoU reward and the role of grounding data, we visualize the evolution of total training rewards and IoU-based rewards throughout the RL phase. These reward dynamics serve as direct indicators of the model's progress in acquiring the \emph{thinking-with-videos} capability.

\begin{figure}[!ht]
  \centering
  \begin{subfigure}{0.45\columnwidth}
    \centering
    \includegraphics[width=\linewidth]{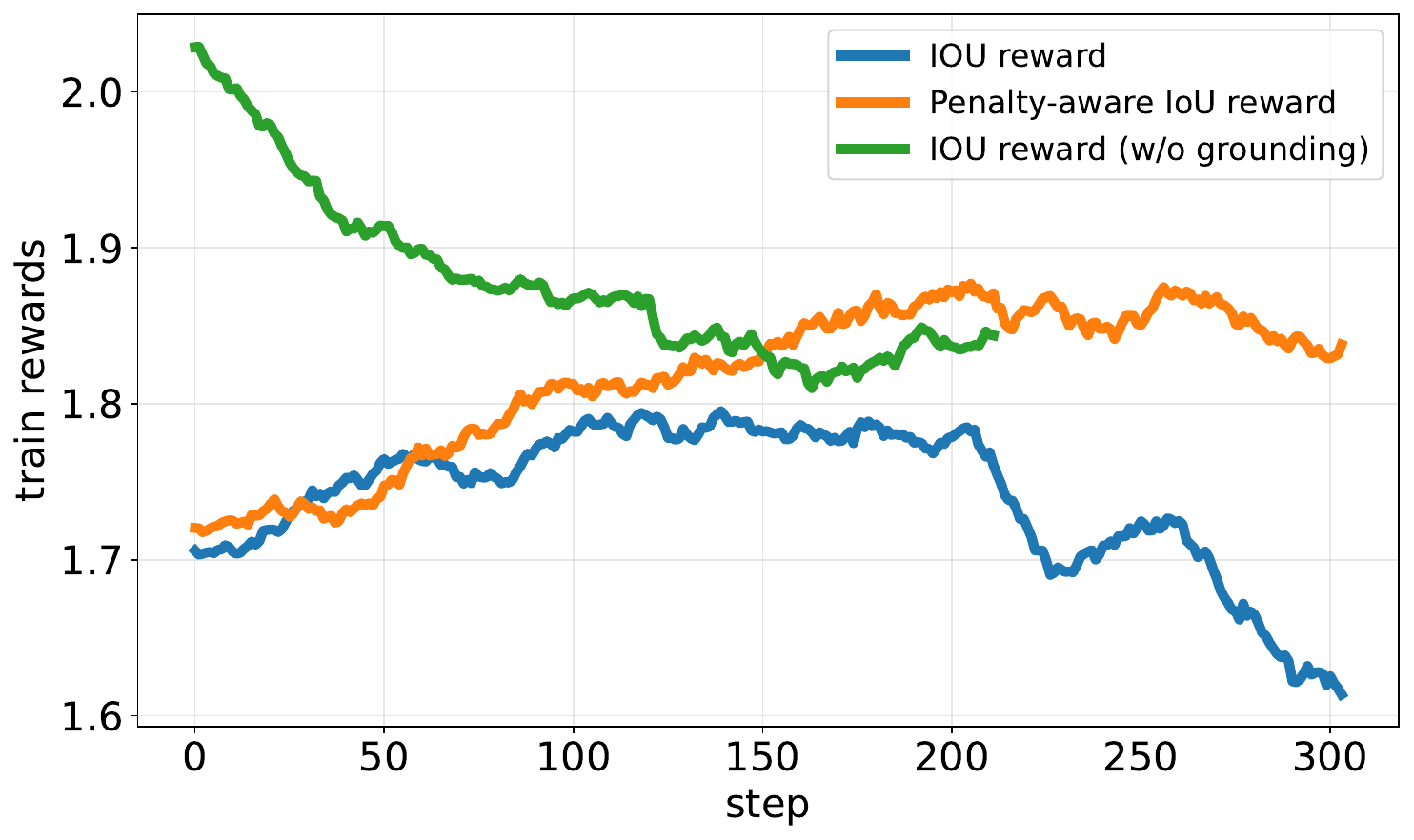}
    \caption{Total training rewards.}
    \label{fig:training_reward}
  \end{subfigure}
  \hspace{0.05\columnwidth}
  \begin{subfigure}{0.45\columnwidth}
    \centering
    \includegraphics[width=\linewidth]{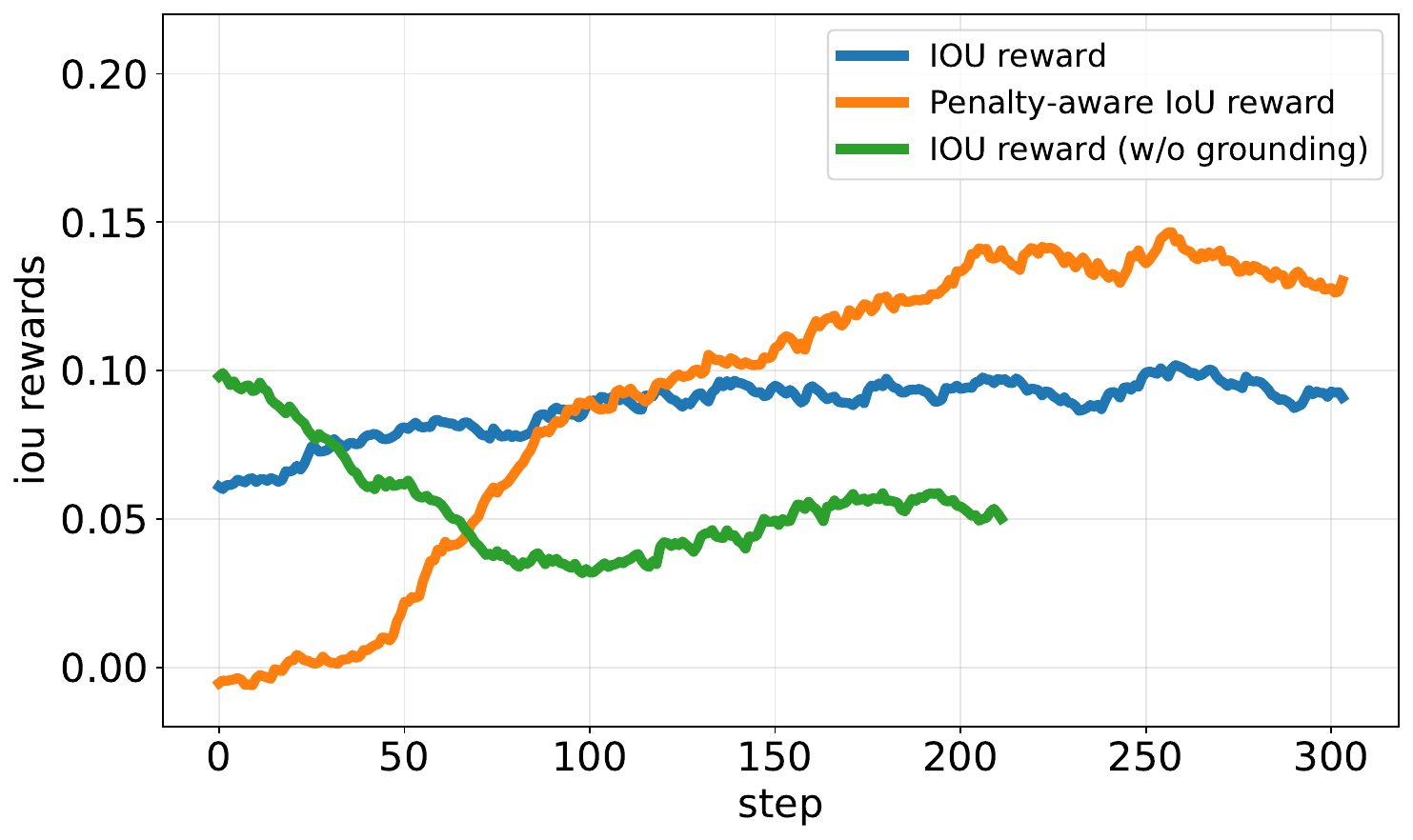}
    \caption{IoU rewards.}
    \label{fig:iou_reward}
  \end{subfigure}
  \caption{Trend of training rewards and IoU rewards.}
  \vspace{-10pt}
  \label{fig:reward}
\end{figure}

As illustrated in Fig.~\ref{fig:training_reward} and Fig.~\ref{fig:iou_reward}, when using the penalty-aware reward, both the total training reward and IoU reward exhibit a consistently stable upward trend. This indicates that model is effectively learning temporal grounding and multi-step reasoning capabilities over the course of RL.
In contrast, using a standard IoU reward (i.e., without penalty) results in a relatively flat IoU reward curve and even a decline in the total training reward. This suggests that the model struggles to reliably localize relevant video segments under this setting, leading to diminished learning effectiveness. Moreover, we observe a higher rate of format collapses, which not only degrade reasoning accuracy but also negatively affect performance on downstream tasks.
In the third configuration, where grounding data is entirely omitted, the training reward initially appear relatively high following SFT. This is likely due to the model's prior learning from pure QA data, which focuses more on general reasoning. However, both the total training reward and the IoU reward deteriorate as training progresses. This decline is likely caused by the lack of grounding supervision, which impairs the model's ability to learn accurate localization behaviors during RL. Consequently, its overall \emph{thinking-with-videos} capacity is undermined.

\section{Open Source Declaration}
This work uses the projects \textit{ms-swift} and \textit{vLLM}, released under the Apache License 2.0.
We strictly adhere to the license requirements. The original projects' copyright notices and full
license texts are available in their official repositories:
\begin{itemize}
  \item \texttt{https://github.com/modelscope/ms-swift}
  \item \texttt{https://github.com/vllm-project/vllm}
\end{itemize}

\newpage
\section{Prompt for Data Construction}

\begin{figure}[!ht]
  \begin{center}
    \centerline{\includegraphics[width=\columnwidth]{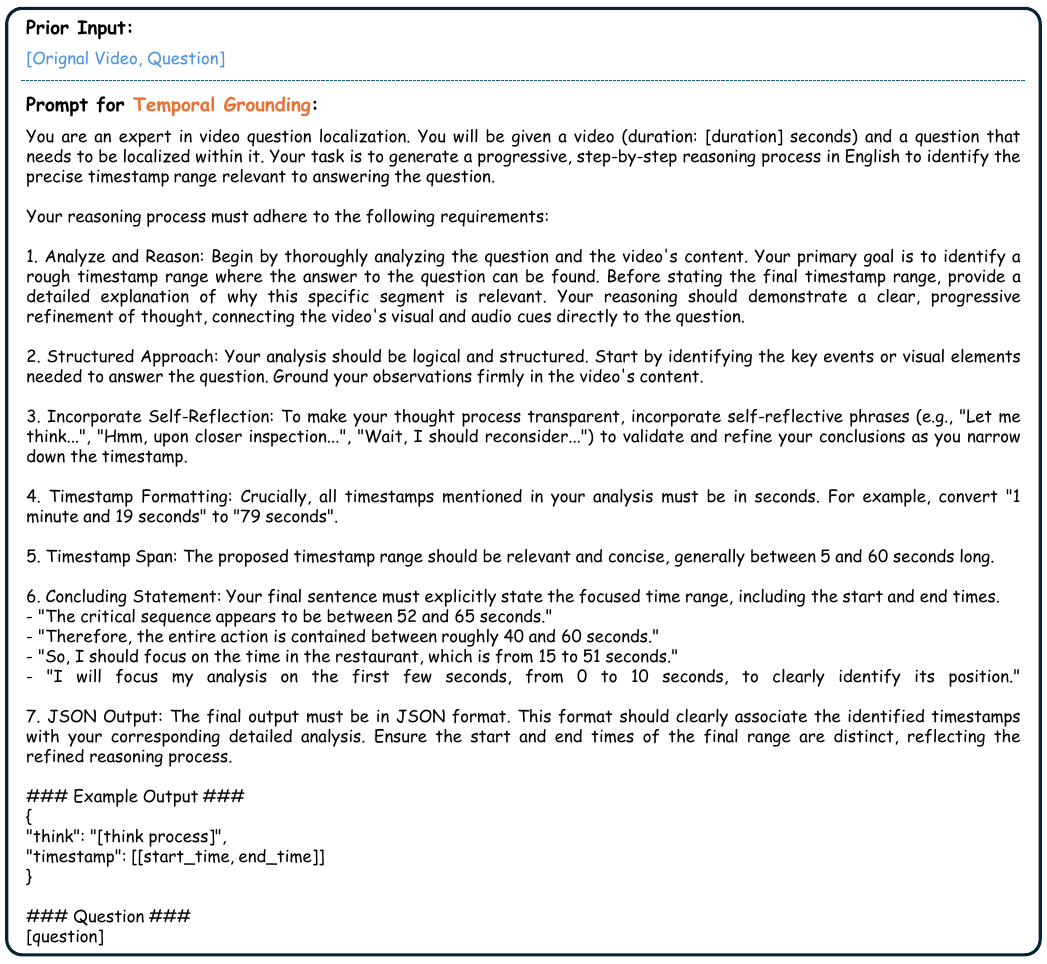}}
    \caption{Prompt for temporal grounding.}
    \label{fig:tg_prompt}
  \end{center}
\end{figure}

\begin{figure}[!ht]
  \begin{center}
    \centerline{\includegraphics[width=\columnwidth]{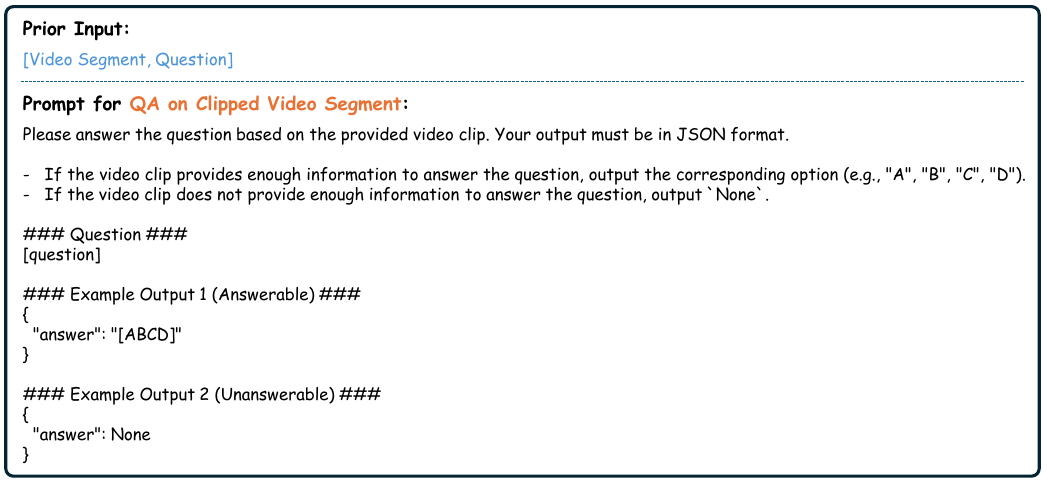}}
    \caption{Prompt for qa on clipped video segment.}
    \label{fig:qa_prompt}
  \end{center}
\end{figure}

\begin{figure}[!ht]
  \begin{center}
    \centerline{\includegraphics[width=\columnwidth]{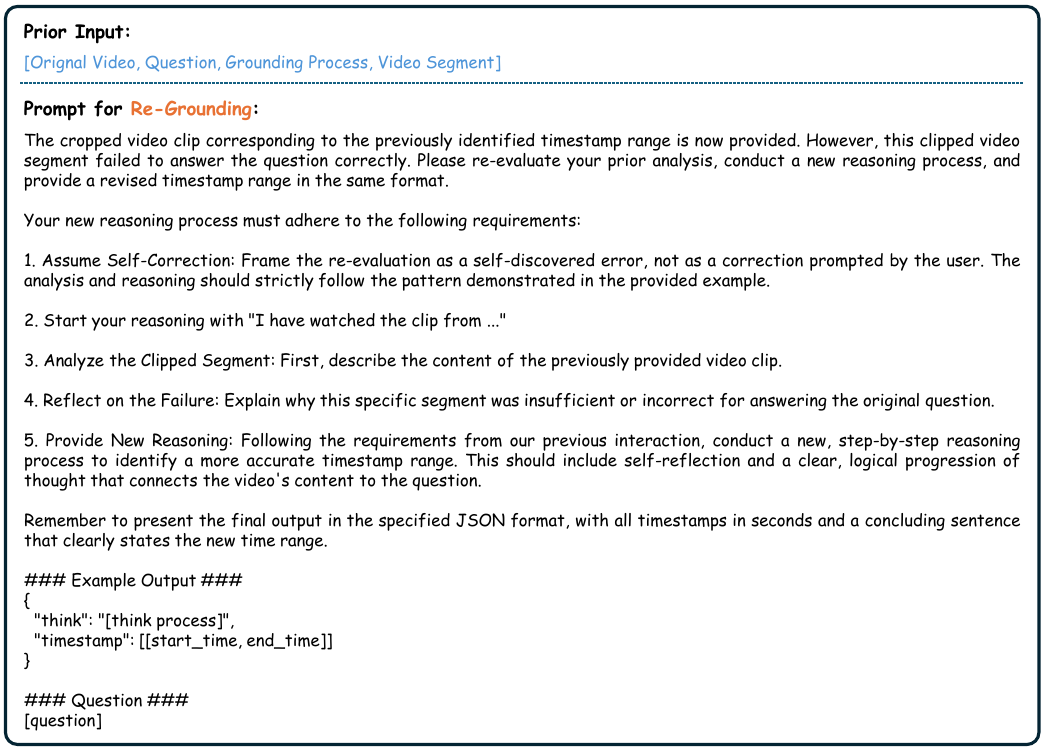}}
    \caption{Prompt for re-grounding.}
    \label{fig:rg_prompt}
  \end{center}
\end{figure}

\begin{figure}[!ht]
  \begin{center}
    \centerline{\includegraphics[width=\columnwidth]{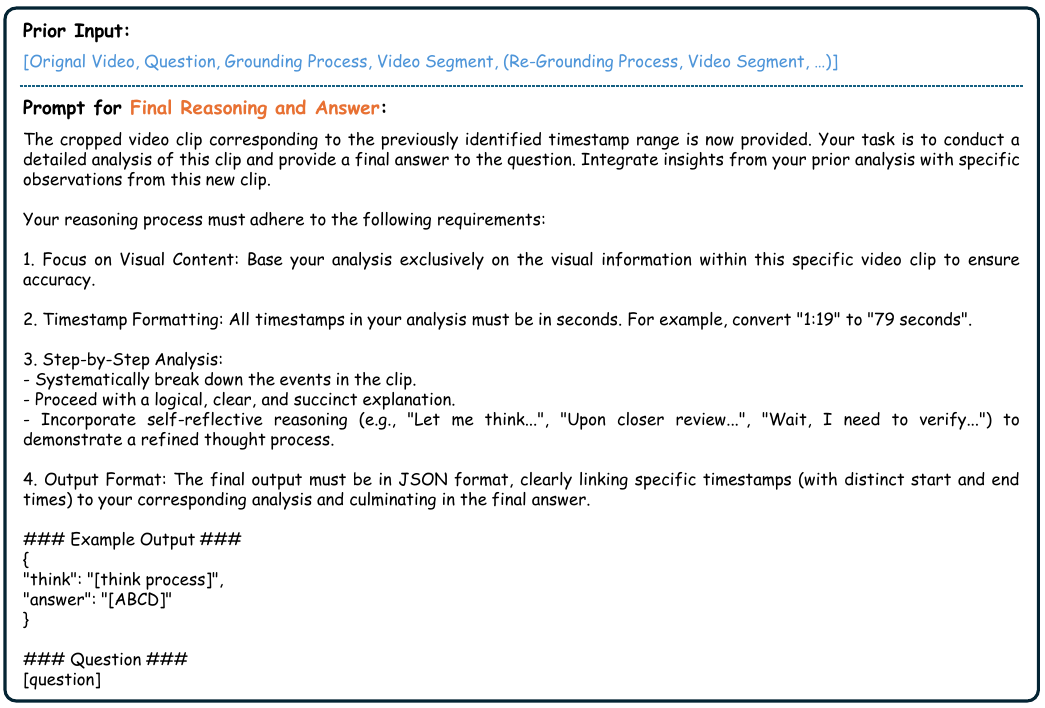}}
    \caption{Prompt for final reasoning and answer.}
    \label{fig:final_prompt}
  \end{center}
\end{figure}

This section presents the prompts used for constructing multi-turn data with tool calls. Specifically, Fig.~\ref{fig:tg_prompt} presents the prompt for temporal grounding, where the model is instructed to identify a relevant video segment given an input question. Fig.~\ref{fig:qa_prompt} shows the prompt used for question answering based solely on the cropped video segment. Fig.~\ref{fig:rg_prompt} illustrates the re-grounding prompt, which is employed when the grounding fails verification, prompting the model to refine its prediction using accumulated context. Finally, Fig.~\ref{fig:final_prompt} depicts the prompt for final reasoning and answer generation, where the model is asked to provide a conclusive answer based on the full context, assuming the correct segment has been identified.

\clearpage
\newpage
\section{Case Study}
\begin{figure}[!h]
  \begin{center}
    \centerline{\includegraphics[width=\columnwidth]{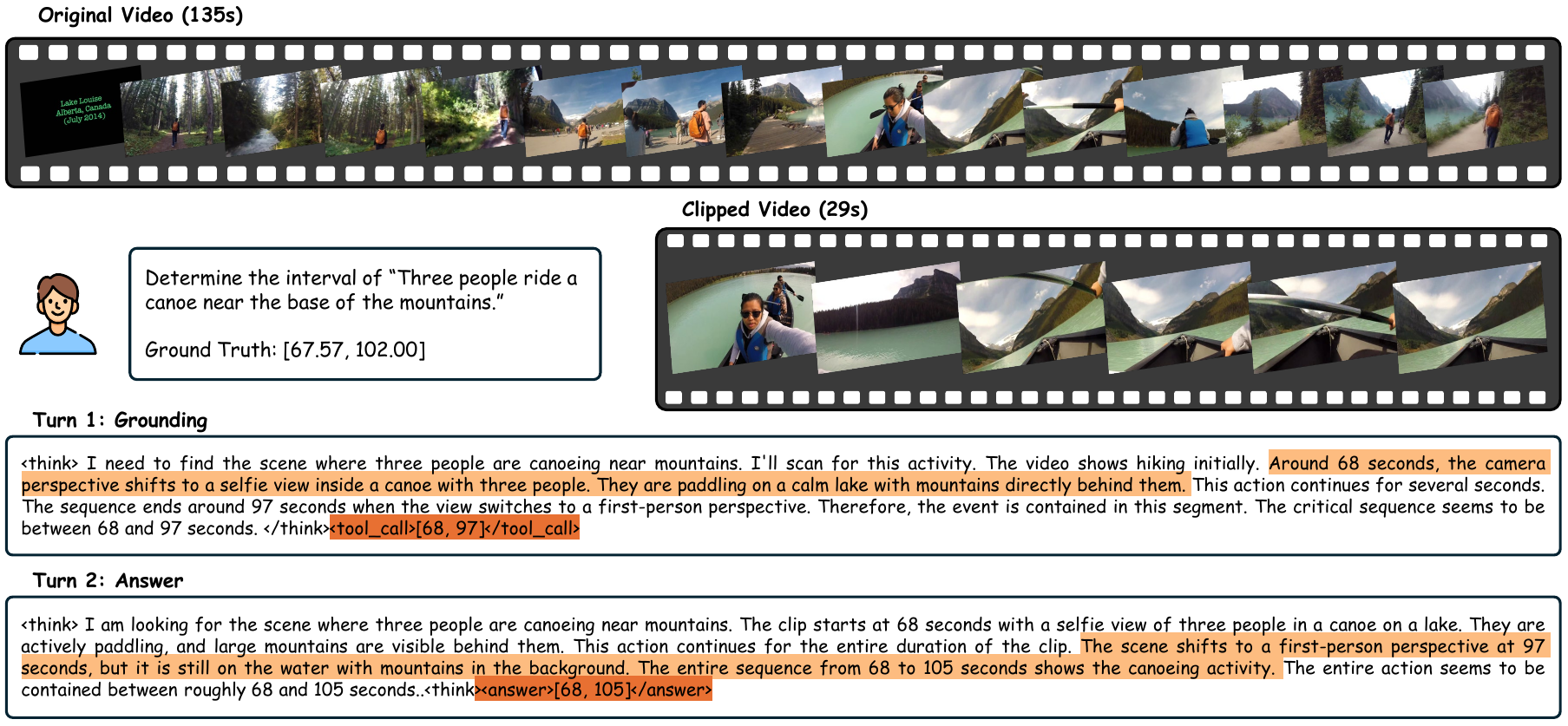}}
    \caption{Grounding task case of \Method.}
    \label{fig:case_1}
  \end{center}
\end{figure}

\begin{figure}[!b]
  \begin{center}
    \centerline{\includegraphics[width=\columnwidth]{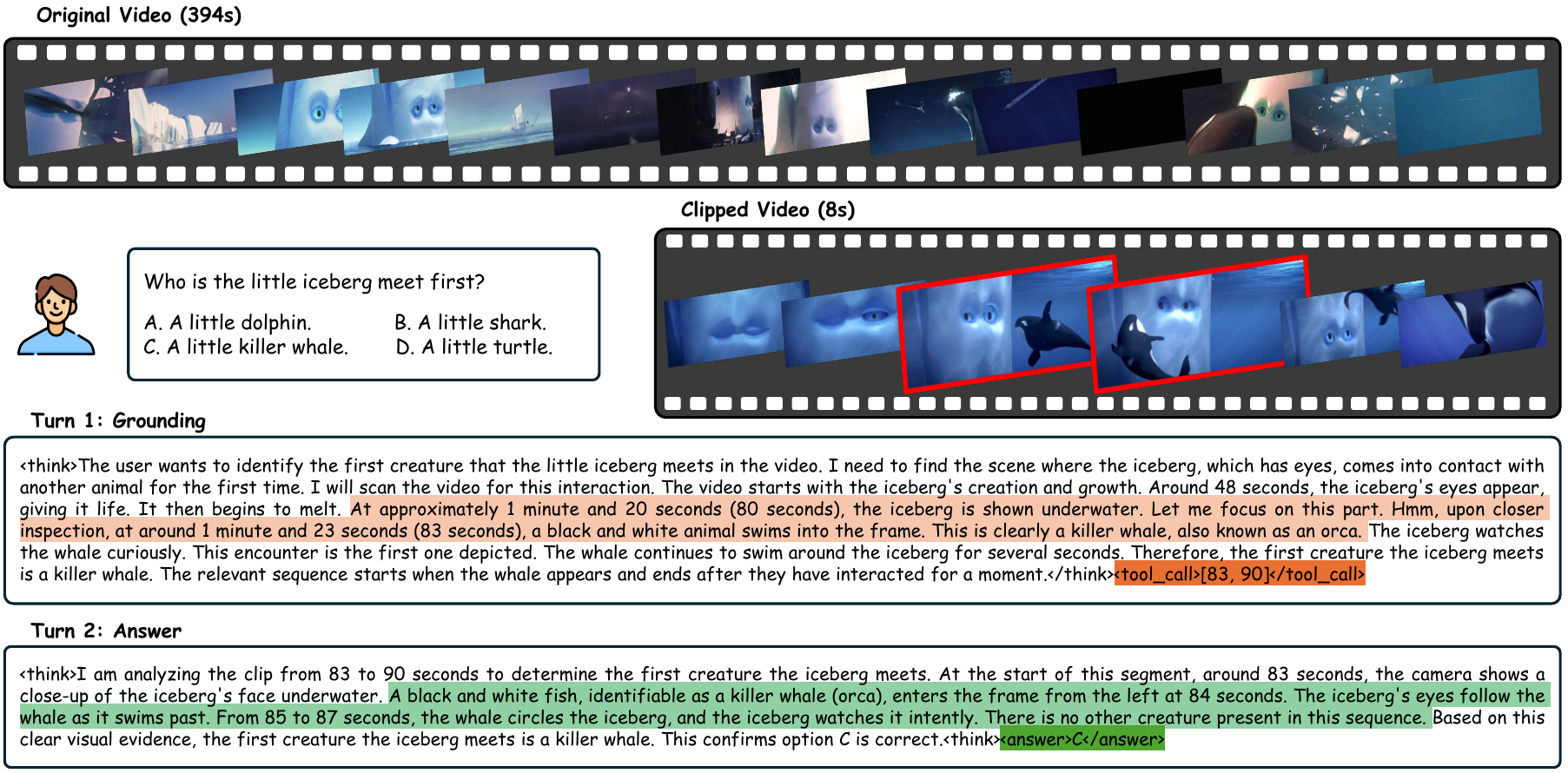}}
    \caption{QA task case 1 of \Method.}
    \label{fig:case_2}
  \end{center}
\end{figure}

\begin{figure}[!t]
  \begin{center}
    \centerline{\includegraphics[width=\columnwidth]{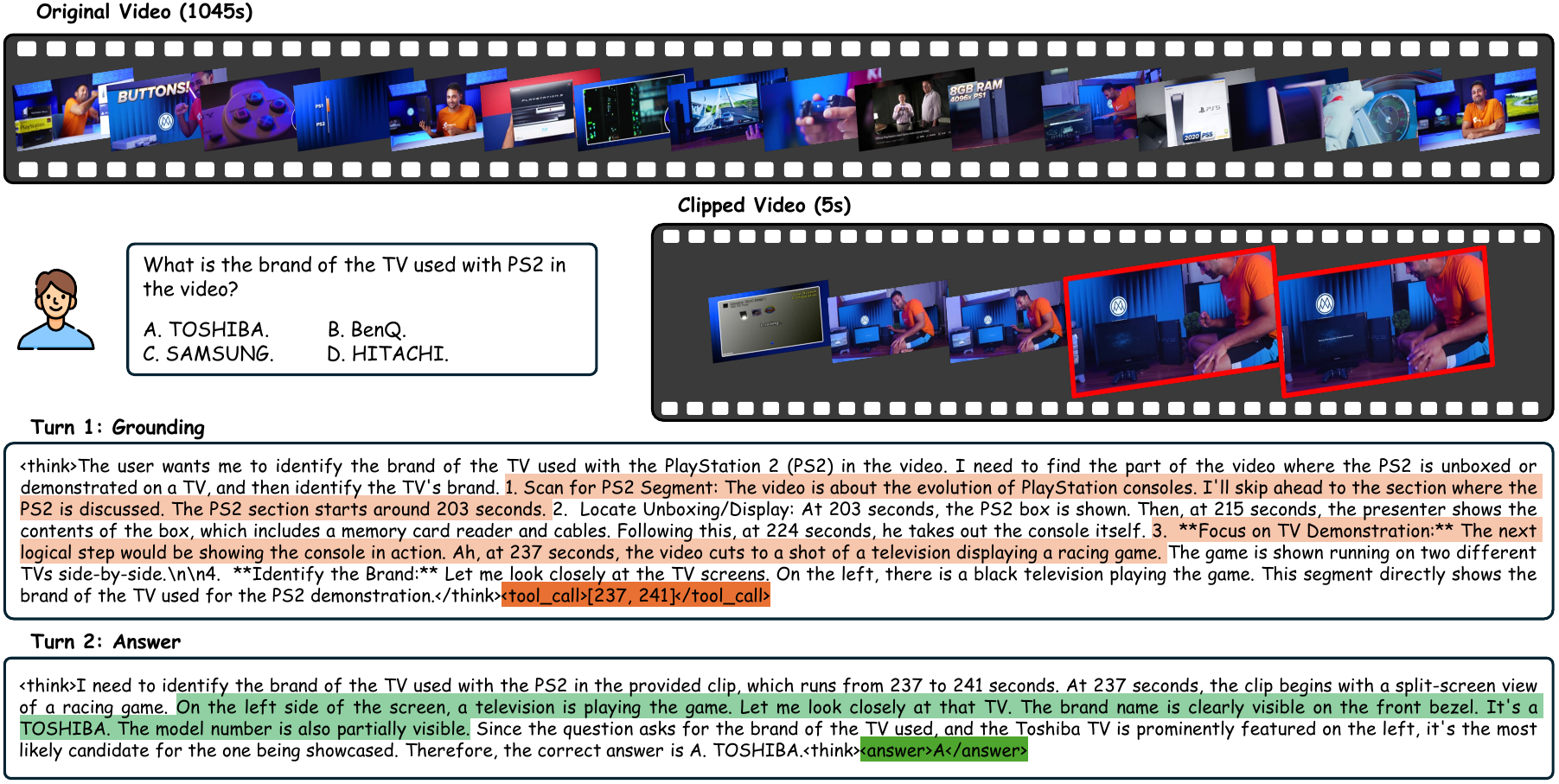}}
    \caption{QA task case 2 of \Method.}
    \label{fig:case_3}
  \end{center}
\end{figure}

\begin{figure}[!b]
  \begin{center}
    \centerline{\includegraphics[width=\columnwidth]{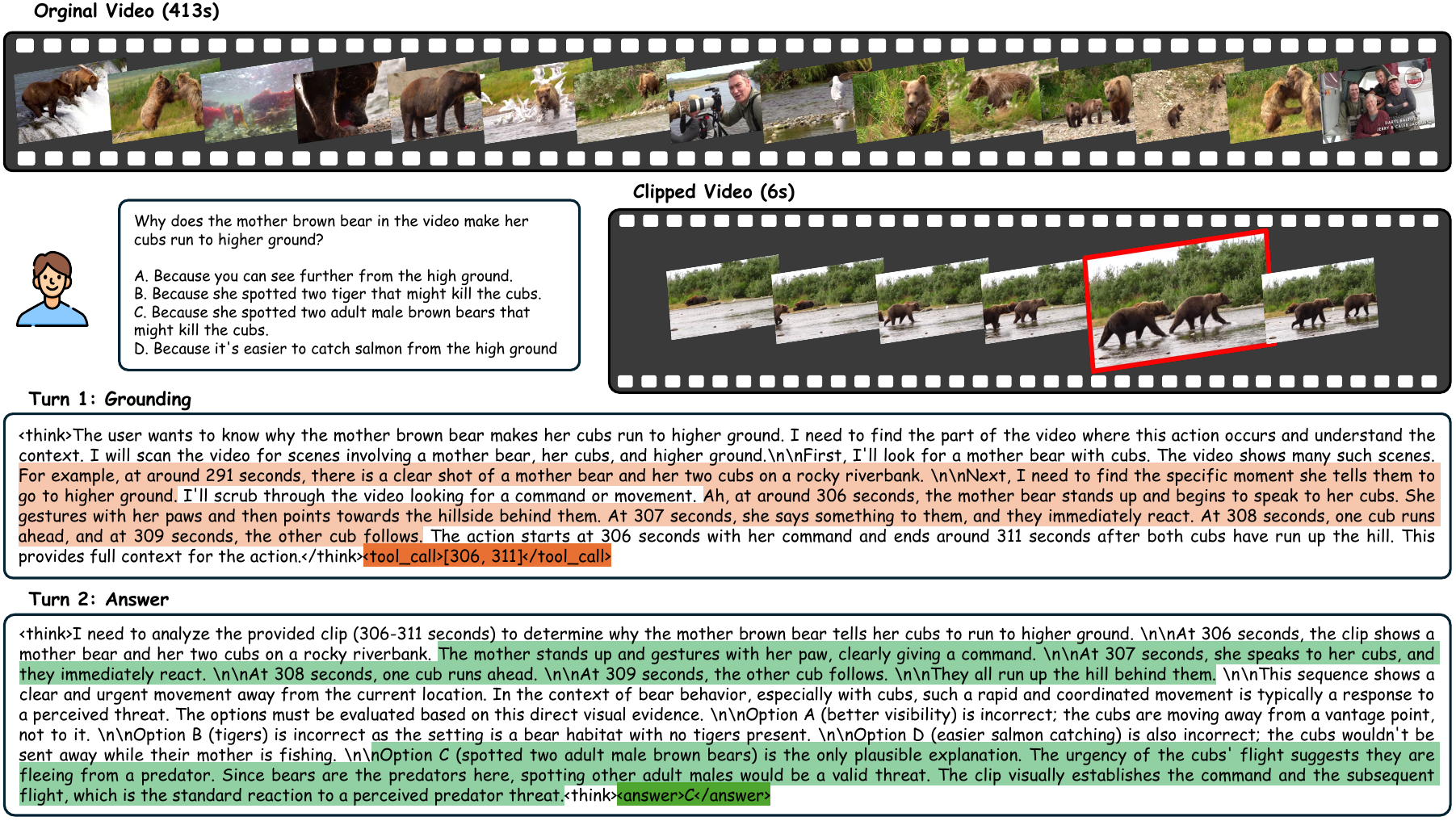}}
    \caption{QA task case 3 of \Method.}
    \label{fig:case_4}
  \end{center}
\end{figure}

\begin{figure}[!t]
  \begin{center}
    \centerline{\includegraphics[width=\columnwidth]{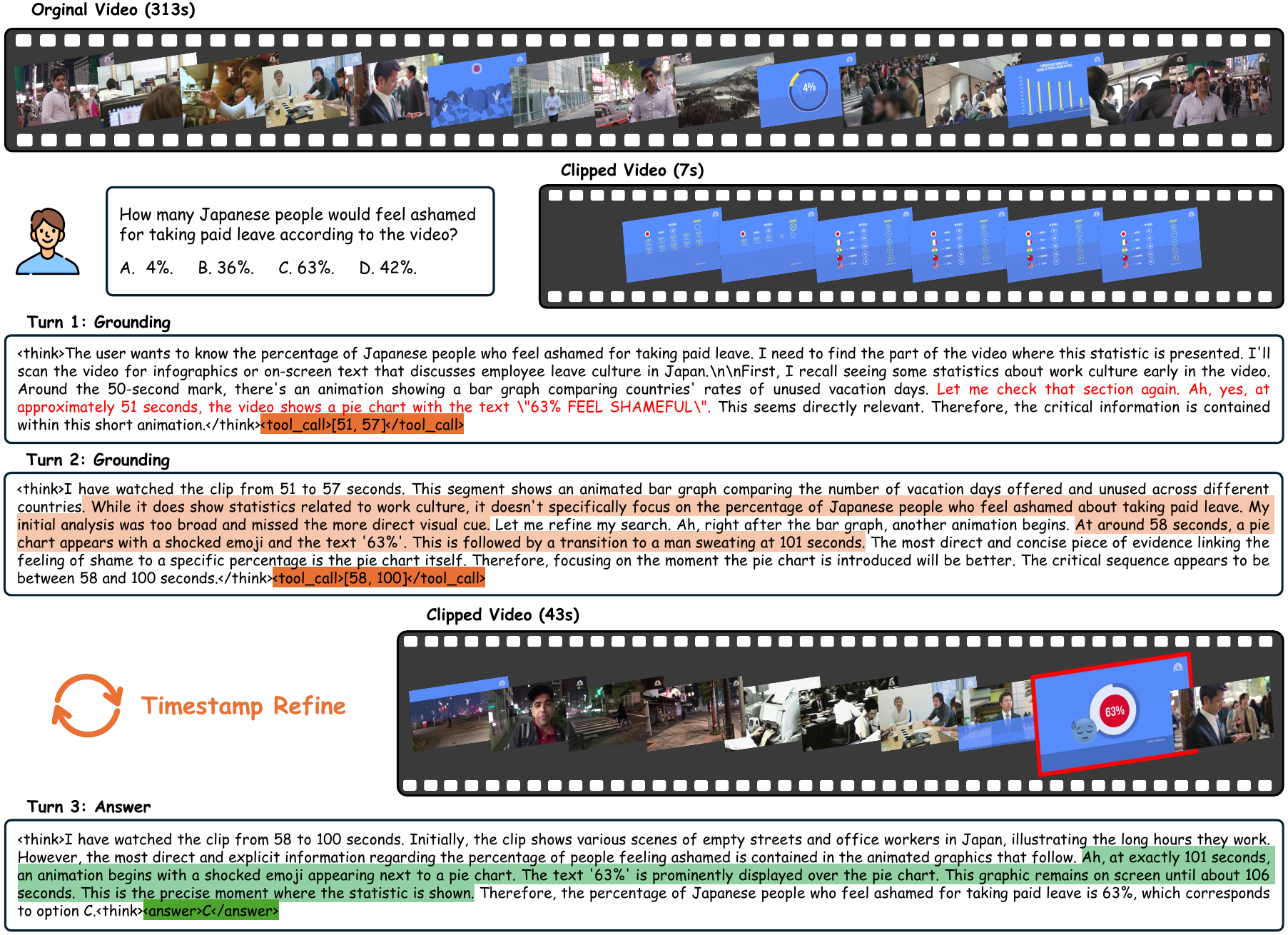}}
    \caption{QA task case 4 of \Method.}
    \label{fig:case_5}
  \end{center}
\end{figure}

\begin{figure}[!t]
  \begin{center}
    \centerline{\includegraphics[width=\columnwidth]{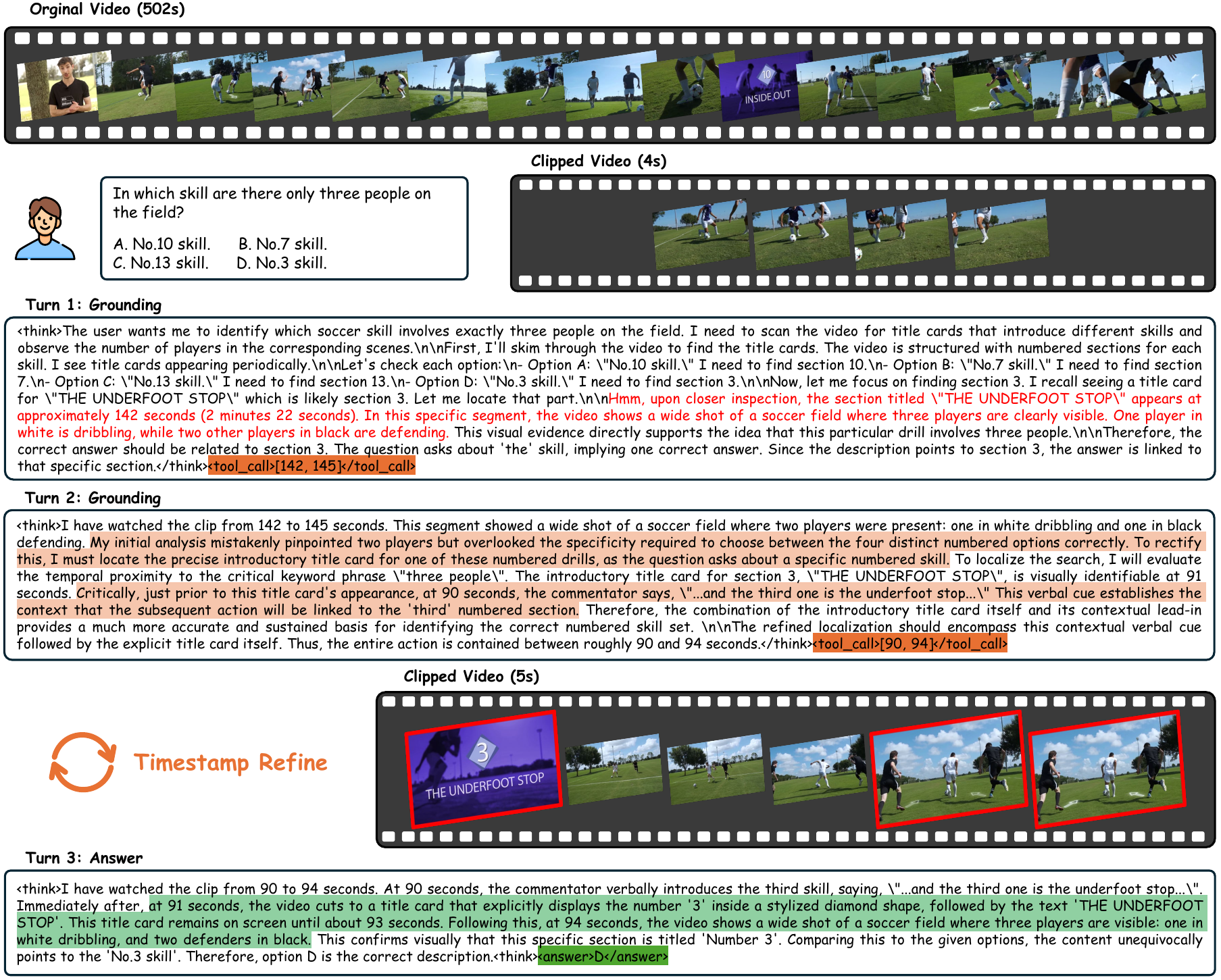}}
    \caption{QA task case 5 of \Method.}
    \label{fig:case_6}
  \end{center}
\end{figure}

To better visualize the behavior of \Method, we present its actual outputs in Fig.~\ref{fig:case_1} to Fig.~\ref{fig:case_6}.

\end{document}